\documentclass[10pt]{article}
\usepackage[accepted]{tmlr}

\usepackage[T1]{fontenc}
\usepackage{amsmath,amsthm,amssymb,graphicx,booktabs}
\usepackage{hyperref}
\usepackage{xurl}
\usepackage{caption}
\usepackage{placeins}
\usepackage{etoc}
\theoremstyle{plain}
\newtheorem{proposition}{Proposition}

\newcommand{\E}{\mathbb{E}}
\newcommand{\Tr}{\operatorname{Tr}}
\newcommand{\R}{\mathbb{R}}
\newcommand{\tW}{\widetilde{W}}
\newcommand{\tw}{\widetilde{w}}
\newcommand{\Sigx}{\Sigma_x}
\newcommand{\Sigeps}{\Sigma_\varepsilon}
\newcommand{\diag}{\operatorname{diag}}
\newcommand{\M}{\mathcal{M}}
\newcommand{\T}{^{\top}}

\title{Learning the identity: a case study of how SGD selects among functional decompositions}

\author{\name Andy Arditi \email arditi.and@northeastern.edu\\
\addr Northeastern University
\AND
\name Weian Xie \\
\addr Massachusetts Institute of Technology
\AND
\name David Bau \\
\addr Northeastern University
\AND
\name Liu Ziyin \email ziyinl@mit.edu \\
\addr Massachusetts Institute of Technology
}

\def\month{12}
\def\year{2026}
\def\openreview{\url{https://openreview.net/forum?id=7Gm2kDkJRe}}

\begin{document}
\etocdepthtag.toc{body}
\maketitle

\begin{abstract}
One might think that learning the identity function with a deep linear residual network is trivial -- the path along residual connections already implements the identity, and so the network need only drive its weights to zero.
However, this zero-weight solution is just one point on an entire manifold of population-loss minimizers, each corresponding to a different decomposition of the identity across the network's layers.
Although the population loss does not distinguish among these solutions, stochastic gradient descent (SGD) reproducibly favors particular ones.
For instance, under anisotropic label noise, the learned layers exhibit a noise-dependent spectrum; even with weight decay, SGD does not generally recover the zero-weight solution.
Changing only the parametrization, while leaving the set of realizable functions unchanged, yields different behavior: factoring each weight matrix as a product of two matrices causes the weights to collapse to zero, even without explicit weight decay.
While perhaps mysterious and unintuitive at first, these phenomena can be understood through the lens of \emph{entropic loss}, which augments the population loss with a term proportional to the expected squared norm of the minibatch gradient \citep{ziyin2025neural}.
On the identity manifold, the population loss is constant, while the entropic term distinguishes among these decompositions.
We characterize its minimizers analytically and use them to derive predictions for the structure of solutions favored by SGD.
Under isotropic label noise, these minimizers are orthogonal factorizations; under anisotropic noise, the first layer's singular values are the fourth roots of the eigenvalues of the noise covariance, and weight decay compresses this spectrum toward one without recovering the zero-weight solution.
In deeper networks, this noise-dependent spectral structure is confined to the first and last layers.
Under the factored parametrization, the entropic term is minimized only at the zero-weight solution.
Networks trained with SGD closely match these predictions.
Overall, the identity learning task studied here serves as a clean and simple case study of how the lens of entropic loss can clarify why SGD favors particular decompositions of the same input--output function.\begingroup
\renewcommand{\thefootnote}{\fnsymbol{footnote}}\footnote[1]{Code available at \url{https://github.com/andyrdt/identity-learning}.}
\endgroup
\end{abstract}

\section{Introduction}\label{sec:intro}

In this work, we study a simple learning problem: fitting a linear residual network to noisy
examples of the identity mapping. At first glance, there seems to be little to explain: the
network can easily represent the identity by setting its weights to zero, so that each layer
passes its input along unchanged via its residual connection. In fact,
\citet{he2016deep} explicitly describe the case of learning the identity mapping as a
motivating example for the use of residual connections.\footnote{``We hypothesize that it is
easier to optimize the residual mapping than to optimize the original, unreferenced mapping.
To the extreme, if an identity mapping were optimal, it would be easier to push the residual
to zero than to fit an identity mapping by a stack of nonlinear layers.''
\citep{he2016deep}} More specifically, we consider a two-layer linear residual network
$f(x)=(I+W_2)(I+W_1)x$ trained with stochastic gradient descent (SGD) on pairs $(x,y)$ with
$y=x+\varepsilon$, where $\varepsilon$ is independent, mean-zero label noise. In this
setting, the population-optimal predictor is the identity, and the network can represent it
simply by setting both weight matrices to zero.

This zero-weight solution is not unique, however. Writing $\tW_\ell:=I+W_\ell$, every factorization $\tW_2\tW_1=I$ -- every \emph{decomposition} of the identity into two effective layers -- achieves the same minimum population loss.
These factorizations form an entire manifold, an \emph{``identity manifold''}, on which the zero-weight configuration is just a single point.
Although the population loss does not distinguish among these solutions, SGD reproducibly favors particular ones.
What determines which decompositions it favors?

This is a simple instance of a broader ambiguity in deep networks: the same input--output function can be realized through many different decompositions across intermediate layers.
The identity task studied here isolates this ambiguity in a setting where the possible decompositions can be characterized analytically.

SGD differs from gradient flow on the population loss in two ways: it takes finite steps and uses noisy minibatch gradients.
For deterministic gradient descent, modified-loss analysis accounts for finite step size through a leading correction proportional to the squared gradient norm \citep{barrett2021implicit}.
Applying this calculation to a fixed minibatch and averaging the resulting correction over minibatches gives the \emph{entropic loss}: the population loss plus a term proportional to the expected squared norm of the minibatch gradient \citep{ziyin2025neural}.
Prior work has used this perspective to derive experimentally supported predictions about the structure of learned solutions \citep{ziyin2025neural}.
Here, we use it to derive and test predictions for solution selection in the identity learning setting.

On the identity manifold, the population loss is constant and cannot distinguish among factorizations.
The entropic term, however, generally varies across them.
We derive this term in closed form and characterize its minimizers, obtaining predictions for how label-noise covariance and network parametrization affect the solutions favored by SGD.
While our analysis does not prove that SGD converges to these minimizers, we find that the predicted solution structures closely match those learned by SGD in the main settings studied here.

Specifically, minimizing the entropic term on the identity manifold yields the following predictions.
Under isotropic label noise, the entropic term is minimized by orthogonal factorizations of the identity.
Under anisotropic noise, the first layer's Gram matrix equals the square root of the noise covariance, implying that its singular values are the fourth roots of the noise-covariance eigenvalues.
In deeper networks, this effect of noise anisotropy is confined to the first and last layers, while interior layers are scaled orthogonal matrices.
Reparameterizing the network by factoring each weight matrix as a product of two matrices causes the weights to collapse to zero, even without weight decay.

Overall, our work highlights the explanatory power of examining learning through the lens of entropic loss, elucidating how SGD selects among decompositions of the same input--output function.

\begin{figure}[t]
    \centering
    \includegraphics[width=0.95\linewidth]{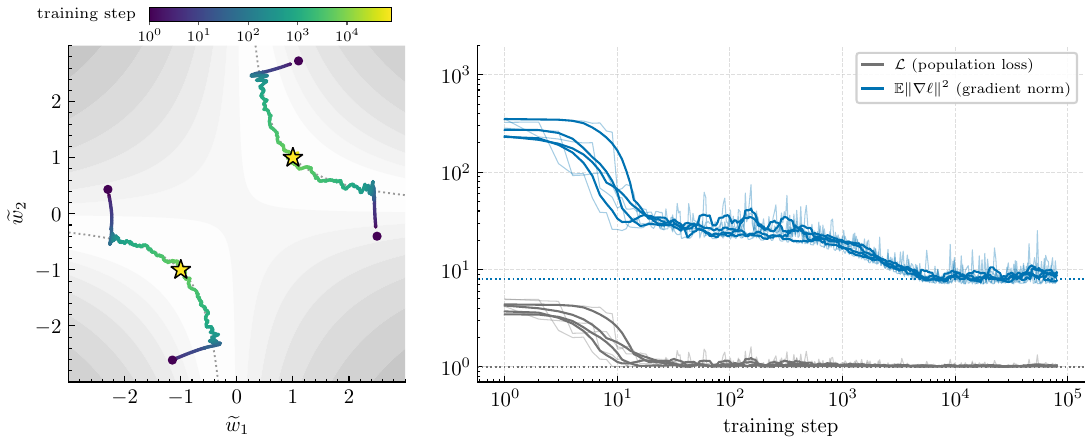}
    \caption{\textbf{The one-dimensional case.} SGD quickly reaches the identity manifold $\tw_2\tw_1=1$, where the population loss is minimized. It then drifts along the manifold toward one of the two factorizations, $\tw_1=\tw_2=1$ or $\tw_1=\tw_2=-1$, that minimize the expected squared gradient norm.
    \emph{Left:} four SGD trajectories through parameter space, colored by training step. Population loss is indicated by shading, and the identity manifold is represented by the dotted hyperbola.
    \emph{Right:} population loss $\mathcal{L}$ and the expected squared gradient norm $\mathbb{E}\|\nabla\ell\|^2$ as functions of training step; dotted lines mark their respective minima on the identity manifold.
    All runs use step size $\eta=0.01$ and batch size $B=1$.}
    \label{fig:oned}
\end{figure}

\section{Setup}\label{sec:setup}

\subsection{Learning the identity function with a linear residual network}\label{sec:model}

To start, we study a two-layer linear residual network:
\[
    f(x)=(I+W_2)(I+W_1)x,
\]
where $W_1,W_2\in\R^{d\times d}$ are learned weight matrices and $I$ is the identity matrix.
We refer to $W_\ell$ as the \emph{residual weight matrix} and $I+W_\ell$ as the \emph{effective layer}.
We use $\widetilde{(\cdot)}$ to denote addition of the identity matrix, writing $\tW_\ell:=I+W_\ell$, so that the network can be written compactly as $f(x)=\tW_2\tW_1x$.

Note that when $W_1=W_2=0$, the network computes the identity function, $f(x)=x$; we call this arrangement of weights the \emph{zero-weight solution}.

We train the network on a noisy identity task: each label $y$ is set to its corresponding
input $x$ plus independent additive noise $\varepsilon$,
\[
    y=x+\varepsilon, \qquad x\sim\mathcal N(0,\Sigx),
    \qquad \varepsilon\sim\mathcal N(0,\Sigeps).
\]
Training uses online SGD: at every step, a fresh minibatch of $B$ samples is drawn from
this distribution.

We assume isotropic inputs, with covariance $\Sigx=I$, and take the label noise covariance to be diagonal, $\Sigeps=\diag(\lambda_1,\dots,\lambda_d)$, with $\lambda_i>0$.\footnote{With isotropic inputs, we can consider diagonal label noise covariance without loss of generality; see Appendix~\ref{app:diag}.}
We normalize the noise covariance by setting $\Tr\Sigeps=\sum_i\lambda_i=d$, so that the average noise variance per coordinate is one.

Since the label noise has mean zero, the optimal predictor is the identity, with irreducible error $\Tr\Sigeps$.
Taking the expectation of the squared error over the data distribution gives the population loss:
\[
    \mathcal L(W_1,W_2) = \E\|y-f(x)\|^2 = \|\tW_2\tW_1 - I\|_F^2 + \Tr\Sigeps .
\]
See Appendix~\ref{app:population-loss} for the derivation.
The second term is constant, and the first vanishes exactly when $\tW_2\tW_1=I$. We call
the resulting set of global minimizers the \emph{identity manifold}:
\[
    \M := \{(W_1,W_2) : \tW_2\tW_1 = I\}.
\]
This set contains infinitely many implementations of the same identity function: for any
invertible matrix $A$, the choice $\tW_1=A$ and $\tW_2=A^{-1}$ lies in $\M$. The zero-weight
solution corresponds to just a single point $A=I$. Since every point in $\M$ has the same population
loss, the population loss alone gives no reason to prefer the zero-weight implementation
over any other factorization of the identity.

\subsection{The one-dimensional case}\label{sec:oned}
Let us first consider the one-dimensional case, where $x,\varepsilon\sim\mathcal N(0,1)$
and $w_1,w_2\in\R$, writing $\tw_\ell:=1+w_\ell$.
The population loss is $\mathcal L=(\tw_2\tw_1-1)^2+1$, minimized on the hyperbola
$\tw_2\tw_1=1$.
Gradient flow on $\mathcal L$ would descend onto this hyperbola and stop wherever it lands.
SGD behaves differently (Figure~\ref{fig:oned}). Each run falls quickly onto the hyperbola
but does not stop there. It drifts slowly along the curve and settles at one of two points:
the zero-weight solution, $\tw_1=\tw_2=1$, or its sign-flipped counterpart,
$\tw_1=\tw_2=-1$, depending on the initialization.\footnote{Appendix~\ref{app:detsector} discusses how initialization can affect the learned factorizations, including in higher dimensional settings.}
This raises the question of why SGD favors these particular points when the population loss does not distinguish among points on the identity manifold.

A clue comes from examining how the stochastic gradients vary along the identity manifold.
Consider the per-sample loss $\ell=(y-\tw_2\tw_1x)^2$.
Differentiating with respect to the weights and then evaluating
on the identity manifold, where $y-\tw_2\tw_1x=\varepsilon$, gives
\[
    \frac{\partial\ell}{\partial w_1}=-2\varepsilon\,\tw_2x,
    \qquad
    \frac{\partial\ell}{\partial w_2}=-2\varepsilon\,\tw_1x.
\]
Using $\E[\varepsilon^2x^2]=1$ and $\tw_2=\tw_1^{-1}$ on the identity manifold,
\[
    \E\|\nabla\ell\|^2
    =4(\tw_1^2+\tw_2^2)
    =4(\tw_1^2+\tw_1^{-2}).
\]
Since $\tw_1^2+\tw_1^{-2}\geq2$, with equality exactly when $\tw_1=\pm1$, the expected squared gradient norm is minimized at $\tw_1=\tw_2=1$ and $\tw_1=\tw_2=-1$.
These are precisely the two factorizations favored by SGD.

\section{The entropic loss}\label{sec:entropic}

The one-dimensional case in Section~\ref{sec:oned} suggests a connection between the solutions favored by finite-step SGD and the expected squared gradient norm.
The \emph{entropic loss}, which augments the population loss with a term proportional to the expected squared norm of the minibatch gradient, provides a framework for understanding this connection.
We now review its construction before specializing it to the identity task.\footnote{See Appendix~\ref{app:entropic} for more detailed derivations.}

Let $\theta$ denote the model parameters.
Relative to gradient flow, SGD differs in two ways: it takes finite steps, and estimates the gradient from a random minibatch.
We consider these effects in turn.

First consider full-batch gradient descent. Gradient flow follows a continuous path
$\dot\theta=-\nabla\mathcal L(\theta)$, whereas gradient descent takes finite steps
$\theta^{(t+1)}\leftarrow\theta^{(t)}-\eta\nabla\mathcal L(\theta^{(t)})$.
Writing $g:=\nabla\mathcal L(\theta)$ and $H:=\nabla^2\mathcal L(\theta)$, gradient flow
run for time $\eta$ gives
\begin{equation}\label{eq:gradient-flow}
    \theta(\eta)
    =\theta-\eta g+\frac{\eta^2}{2}Hg+O(\eta^3),
\end{equation}
while a gradient-descent step contains no corresponding second-order term (or higher order terms).
Thus, after a time interval of length $\eta$, gradient flow and one gradient-descent step differ by $O(\eta^2)$.

Backward error analysis \citep{hairer2006geometric} asks whether a nearby continuous
flow can match the discrete update more closely.
Applied to gradient descent, it gives a modified loss \citep{barrett2021implicit}
\begin{equation}\label{eq:modified-loss}
    \mathcal L_\eta(\theta)
    =\mathcal L(\theta)+\frac{\eta}{4}\|\nabla\mathcal L(\theta)\|^2+O(\eta^2).
\end{equation}
The gradient of the added term is $\frac{\eta}{2}Hg$, so its contribution to the
modified flow cancels the second-order displacement in Equation~\ref{eq:gradient-flow}.
After time $\eta$, gradient flow on $\mathcal L_\eta$ therefore differs from one
gradient-descent step on $\mathcal L$ only by $O(\eta^3)$.

SGD replaces the population gradient with a minibatch gradient. Let $\mathcal B$ be a minibatch consisting of $B$ independent examples, and define $\ell_{\mathcal B}=\frac{1}{B}\sum_{b\in\mathcal B}\ell_b$, where $\ell_b$ is the loss on example $b$. For each fixed minibatch, the modified-loss calculation in Equation~\ref{eq:modified-loss} gives
\begin{equation}
 \ell_{\mathcal B,\eta}(\theta)
 =\ell_{\mathcal B}(\theta)
 +\frac{\eta}{4}\|\nabla\ell_{\mathcal B}(\theta)\|^2+O(\eta^2).
\end{equation}
Following \citet{ziyin2025neural}, we average the first-order correction over the minibatch distribution and define the first-order \emph{entropic loss}
\begin{equation}\label{eq:Lent}
 F_\eta(\theta):=\mathcal L(\theta)+S(\theta),
 \qquad S(\theta):=\frac{\eta}{4}\E_{\mathcal B}
 \|\nabla\ell_{\mathcal B}(\theta)\|^2.
\end{equation}
We refer to $S$ as the \emph{entropic term}, and its corresponding force $-\nabla S$ as the \emph{entropic force}.
Prior work has used the entropic loss perspective to derive experimentally supported predictions about the structure of solutions favored by SGD, beyond what is determined by the training loss alone \citep{ziyin2025neural}.

Here, we apply this perspective to the identity learning setting: we characterize minimizers of the entropic objective on the identity manifold and test whether the resulting predictions describe the solutions learned by SGD.

Note that the minibatch gradient has mean $\nabla\mathcal L$, while averaging $B$ independent examples reduces its variance by a factor of $B$.
The usual bias--variance decomposition therefore gives
\begin{equation}\label{eq:gradient-bias-variance}
    \E_{\mathcal B}\big\|\nabla\ell_{\mathcal B}\big\|^2
    =\|\nabla \mathcal L\|^2
    +\frac{1}{B}\,\E\big\|\nabla\ell-\nabla \mathcal L\big\|^2.
\end{equation}
Multiplying by $\eta/4$, the first term gives the correction already present in full-batch gradient descent, while the second gives the contribution from minibatch noise.

\paragraph{Specialization to the identity task.}
We now return to the identity learning task specified in Section~\ref{sec:model}.
On the identity manifold $\M$, the population gradient $\nabla\mathcal L$ vanishes, so, using Equations~\ref{eq:Lent} and~\ref{eq:gradient-bias-variance}, the entropic term reduces to
\begin{equation}\label{eq:S-on-M}
    S(\theta)=\frac{\eta}{4B}\,\E\|\nabla\ell(\theta)\|^2.
\end{equation}
Thus, up to the constant factor $\eta/(4B)$, the entropic term on the identity manifold is exactly the expected squared gradient norm examined in Section~\ref{sec:oned}.
Because the population loss is constant on $\M$, minimizing the entropic loss on this manifold is equivalent to minimizing $S$.
This yields a prediction for which factorizations SGD favors.
In one dimension, $\E\|\nabla\ell\|^2$ is minimized at $\tw_1=\tw_2=\pm1$, recovering the two factorizations observed in Figure~\ref{fig:oned}.

\paragraph{Weight decay.}
When weight decay is used, the objective additionally includes $\gamma\|\theta\|^2$, where $\|\theta\|^2=\|W_1\|_F^2+\|W_2\|_F^2$.
Applying the same modified-loss construction gives, on the identity manifold,
\begin{equation}\label{eq:Lent-wd-M}
    F_{\eta,\gamma}(\theta)
    =\mathcal L(\theta)
    +\gamma(1+\eta\gamma)\|\theta\|^2
    +S(\theta).
\end{equation}
In all of our experiments $\eta\gamma\leq0.02$, so we generally ignore this small rescaling and use $\mathcal L(\theta)+\gamma\|\theta\|^2+S(\theta)$ as the effective objective.

In the next section, we characterize the minimizers of these effective objectives on the identity manifold and compare the resulting predictions with SGD experiments in higher dimensions ($d>1$), both with and without weight decay.

\section{How the entropic term selects among identity factorizations}\label{sec:select}

We now minimize the entropic loss over identity factorizations to derive predictions for the solutions favored by SGD, and compare these predictions with experiments.
We begin with two-layer linear residual networks (Section~\ref{sec:two-layer-selection}), examining how noise covariance and weight decay affect the predicted solutions (Sections~\ref{sec:iso} and~\ref{sec:aniso}).
We then consider deeper networks (Section~\ref{sec:deep}) and alternative parametrizations (Section~\ref{sec:factored}).

\subsection{Minimizing the entropic term over the identity manifold}\label{sec:two-layer-selection}

Our interest is in how SGD selects among solutions that implement the identity.
We therefore minimize the entropic loss restricted to the identity manifold $\M$, where the population loss is constant and the optimization reduces to minimizing $S$.
This constrained problem is motivated by the small-step-size regime: the population loss is the leading term in $F_\eta=\mathcal L+S$, while $S=O(\eta)$ distinguishes its minimizers.
We use the resulting factorizations as predictions for SGD's learned solutions.\footnote{At finite step size, minimizing the full entropic loss need not give a solution exactly on $\M$.}

Recall the $d$-dimensional setting, in which we train a two-layer linear residual network
\[
    f(x)=\tW_2\tW_1x=(I+W_2)(I+W_1)x,
    \qquad \tW_\ell:=I+W_\ell,
\]
with $W_1,W_2\in\R^{d\times d}$. Inputs are drawn from
$x\sim\mathcal N(0,I_d)$, and labels are
$y=x+\varepsilon$, where
$\varepsilon\sim\mathcal N(0,\Sigeps)$.
We take
$\Sigeps=\diag(\lambda_1,\dots,\lambda_d)$ and normalize
$\Tr\Sigeps=d$.

\paragraph{Computing the entropic term on the identity manifold.}
On the identity manifold $\M$, we have $\tW_2=\tW_1^{-1}$, and so the prediction error for any given example is simply the label noise:
$y-f(x)=\varepsilon$.

Differentiating the per-example loss with respect to the residual weights $W_\ell$ gives
\[
    \nabla_{W_1}\ell=-2(\tW_2\T\varepsilon)x\T,
    \qquad
    \nabla_{W_2}\ell=-2\varepsilon(\tW_1x)\T.
\]
Note that since $\tW_\ell=I+W_\ell$, differentiation with respect to $W_\ell$ and $\tW_\ell$ is equivalent.
Each gradient is an outer product of two independent random vectors.
Using
$\|uv\T\|_F^2=\|u\|^2\|v\|^2$ and
$\E\|Az\|^2=\Tr(A\Sigma_zA\T)$, together with
$\E\|x\|^2=\E\|\varepsilon\|^2=d$, gives
\[
    \E\|\nabla_{W_1}\ell\|_F^2
    =4d\,\Tr(\tW_2\T\Sigeps\tW_2),
    \qquad
    \E\|\nabla_{W_2}\ell\|_F^2
    =4d\,\Tr(\tW_1\tW_1\T).
\]
Summing these two terms, multiplying by $\frac{\eta}{4B}$ as in
Equation~\ref{eq:S-on-M}, and substituting
$\tW_2=\tW_1^{-1}$ yields
\begin{equation}\label{eq:S}
    S(\tW_1)=\frac{\eta d}{B}\left[
    \Tr\!\big(\tW_1\tW_1\T\big)
    +\Tr\!\big(\tW_1^{-\top}\Sigeps\tW_1^{-1}\big)
    \right].
\end{equation}

\paragraph{Reducing the optimization to the Gram matrix.}
Equation~\ref{eq:S} simplifies considerably if we define the Gram matrix $G:=\tW_1\T\tW_1$.
The two traces in Equation~\ref{eq:S} can then be written as
\[
    \Tr(\tW_1\tW_1\T)=\Tr G,
    \qquad
    \Tr(\tW_1^{-\top}\Sigeps\tW_1^{-1})
    =\Tr(\Sigeps G^{-1}),
\]
and so the entropic term depends on $\tW_1$ only through its Gram matrix $G$:
\begin{equation}\label{eq:S-gram}
    S(G)=\frac{\eta d}{B}
    \left[\Tr G+\Tr(\Sigeps G^{-1})\right].
\end{equation}
On the identity manifold, $G=\tW_1\T\tW_1$ is positive definite.
Conversely, every positive-definite $G$ is attainable by taking $\tW_1=G^{1/2}$ and $\tW_2=G^{-1/2}$.
Thus the constrained problem is equivalent to minimizing Equation~\ref{eq:S-gram} over $G\succ0$.

Differentiating with respect to $G$ gives
\[
    \nabla_G S(G)
    =\frac{\eta d}{B}
    \left(I-G^{-1}\Sigeps G^{-1}\right).
\]
Setting the gradient to zero yields $G^{-1}\Sigeps G^{-1}=I$, or equivalently $G^2=\Sigeps$.
The objective in Equation~\ref{eq:S-gram} is strictly convex over $G\succ0$ (Appendix~\ref{app:gram-derivative}), so this stationary point is the unique minimizing Gram matrix.

\paragraph{The resulting weight structure.}
Since $G$ is symmetric positive definite, the equation $G^2=\Sigeps$ has the unique symmetric positive-definite solution $G=\Sigeps^{1/2}$.
Hence the minimizing factorizations satisfy
\begin{equation}\label{eq:gram-balance}
    \tW_1\T\tW_1=\Sigeps^{1/2}.
\end{equation}

Equation~\ref{eq:gram-balance} determines $\tW_1$ up to an orthogonal transformation on the left.
Therefore every minimizing factorization can be written as
\[
    \tW_1=U\Sigeps^{1/4},
    \qquad
    \tW_2=\Sigeps^{-1/4}U\T,
\]
for some orthogonal matrix $U$.

Since $\Sigeps=\diag(\lambda_1,\dots,\lambda_d)$ in our chosen basis, the singular values of $\tW_1$ are
\begin{equation}\label{eq:balance}
    s_i=\lambda_i^{1/4}.
\end{equation}
The right singular vectors of $\tW_1$ can be chosen as eigenvectors of $\Sigeps$; within a repeated-eigenvalue eigenspace, their choice is arbitrary.
The entropic term therefore determines the singular values and their associated input eigenspaces, but leaves an arbitrary orthogonal transformation $U$ of the hidden representation.

\begin{figure}[t]
    \centering
    \includegraphics[width=0.98\linewidth]{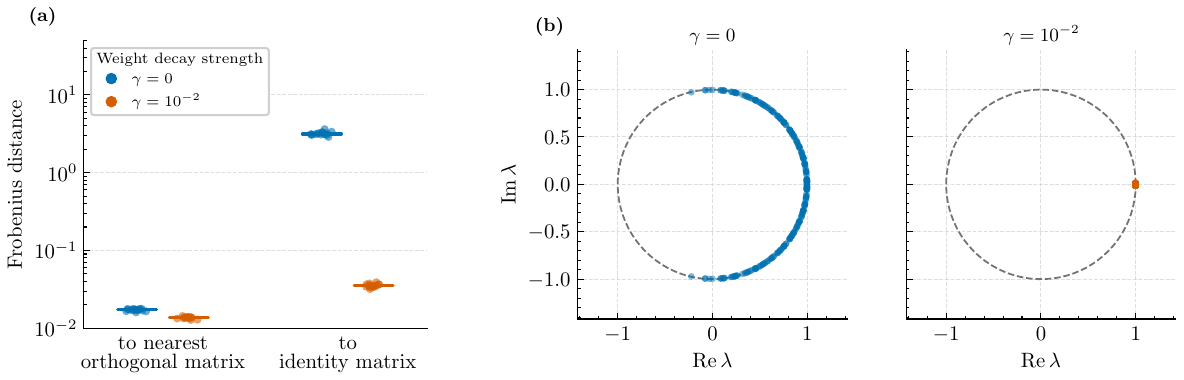}
    \caption{\textbf{Isotropic noise selects orthogonal factorizations, while weight decay favors the identity.}
    \emph{(a)} Frobenius distance of the learned effective layer $\tW_1$ to its nearest orthogonal matrix and to the identity matrix.
    Each point represents one of 16 seeds; horizontal bars show medians.
    With or without weight decay, the learned layers are nearly orthogonal, while weight decay ($\gamma=10^{-2}$) additionally brings them close to the identity.
    \emph{(b)} Eigenvalues of the same learned matrices in the complex plane (dashed: unit circle), with all 16 eigenvalues from each of 16 seeds overlaid in each subplot.
    Without weight decay, the eigenvalues spread along an arc near the unit circle; with weight decay, they concentrate near $+1$.
    Here $d=16$, $\eta=0.02$, $B=64$, and $(W_\ell)_{ij}\sim\mathcal N(0,1/d)$.
    We train for $2\times10^6$ steps and average the weights over the final $1\%$.}
    \label{fig:eig}
\end{figure}

\subsection{Isotropic noise selects orthogonal solutions}\label{sec:iso}

With isotropic noise, $\Sigeps=I$, Equation~\ref{eq:gram-balance} reduces to $\tW_1\T\tW_1=I$, and thus $\tW_1$ is orthogonal.
The minimizing factorizations therefore have the form $\tW_1=U$ and $\tW_2=U\T$, for any orthogonal matrix $U$.

Figure~\ref{fig:eig}(a) supports this prediction empirically: without weight decay, the learned
effective layers lie close to the orthogonal group.
Figure~\ref{fig:eig}(b) visualizes the remaining orthogonal freedom through the eigenvalues of the
learned matrices, which lie near the unit circle.\footnote{Their concentration in the right half-plane reflects the initialization around $I$; see Appendix~\ref{app:detsector}.}

\paragraph{Adding weight decay.}
Weight decay breaks the indifference among orthogonal factorizations.
The zero-weight solution, $\tW_1=\tW_2=I$ (i.e., $W_1=W_2=0$), minimizes the entropic term and uniquely minimizes the weight-decay penalty.
It is therefore the unique minimizer of their sum on the identity manifold.
Figure~\ref{fig:eig}(a) shows the learned effective layers moving toward $I$, while Figure~\ref{fig:eig}(b) shows their eigenvalues concentrating near $+1$.

\begin{figure}[t]
    \centering
    \includegraphics[width=0.98\linewidth]{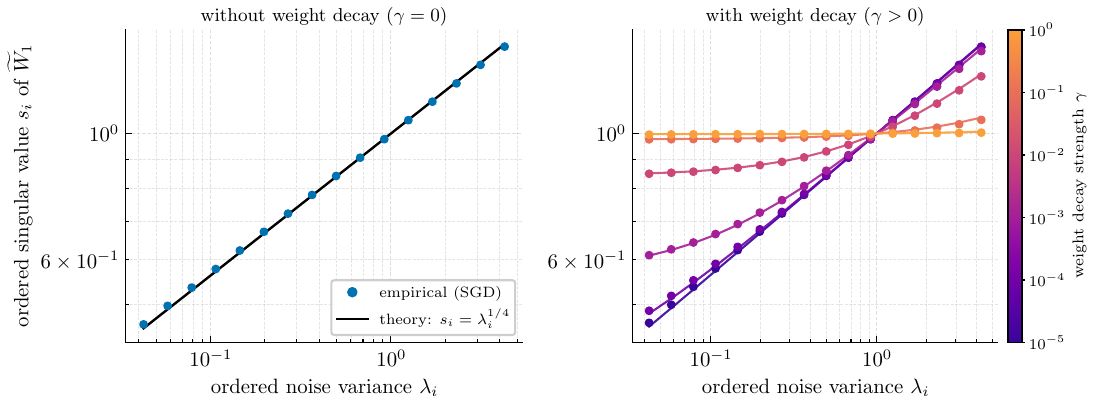}
    \caption{\textbf{Anisotropic noise induces a fourth-root singular-value spectrum, while weight decay pulls the singular values toward one.}
    \emph{Left:} Without weight decay, SGD closely follows the fourth-root law
    $s_i=\lambda_i^{1/4}$.
    Points show mean learned singular values over five runs.
    \emph{Right:} With weight decay, the singular-value spectrum is compressed toward one,
    closely matching theoretical predictions.
    Points show mean learned singular values over five runs, and lines show the corresponding predictions of Equation~\ref{eq:quartic}; colors indicate the weight-decay strength $\gamma$.
    The predictions retain the correction $\gamma\mapsto\gamma(1+\eta\gamma)$ described in Appendix~\ref{app:quartic-numerics}.
    In both panels, singular values are sorted within each run and paired with the ordered noise variances; error bars show one standard deviation and are smaller than the markers.
    Here $d=16$, $\eta=0.005$, and $B=16$.
    Singular values are computed from weights averaged over the final $1\%$ of training.}
    \label{fig:fourthroot}
\end{figure}

\subsection{Anisotropic noise gives the fourth-root law}\label{sec:aniso}

Under anisotropic noise, Equation~\ref{eq:balance} predicts that the singular values of
$\tW_1$ satisfy $s_i=\lambda_i^{1/4}$.
Figure~\ref{fig:fourthroot} (left) tests this fourth-root law by comparing the ordered
singular values learned by SGD with the ordered eigenvalues of the noise covariance.

\paragraph{Adding weight decay.}
The fourth-root law above describes the case $\gamma=0$.
The weight decay penalty alone favors the zero-weight solution, for which every singular value of $\tW_1$ equals one.
Under anisotropic noise, this preference competes with the entropic term.
In the eigenbasis of the noise covariance, a minimizer of the on-manifold objective can be chosen diagonal and positive (Appendix~\ref{app:wd-diagonal}).
We may therefore write $\tW_1=\diag(a_i)$ and $\tW_2=\diag(a_i^{-1})$, where the $a_i>0$ are the singular values of $\tW_1$. Substituting these forms into the entropic term and weight decay penalty gives
\begin{equation}\label{eq:phi}
    S+\gamma\big(\|W_1\|_F^2+\|W_2\|_F^2\big)
    =\sum_i\phi_i(a_i),
\end{equation}
where
\[
    \phi_i(a_i)
    =\frac{\eta d}{B}(a_i^2+\lambda_i a_i^{-2})
    +\gamma\big((a_i-1)^2+(a_i^{-1}-1)^2\big).
\]
Thus each noise mode can be optimized independently.
For mode $i$, the entropic term favors the fourth-root solution
$a_i=\lambda_i^{1/4}$, while weight decay favors $a_i=1$.
The selected value balances these two effects.

Setting $\phi_i'(a_i)=0$ and multiplying by $a_i^3/2$ gives
\begin{equation}\label{eq:quartic}
    \big(\gamma+\tfrac{\eta d}{B}\big)a_i^4
    -\gamma a_i^3+\gamma a_i
    -\big(\gamma+\tfrac{\eta d}{B}\lambda_i\big)=0.
\end{equation}
This equation has a unique positive root, which gives our on-manifold prediction for
the singular value $a_i$.
Figure~\ref{fig:fourthroot} (right) shows that these predictions closely match the singular
values learned by SGD.
For $\lambda_i\neq1$, the predicted singular value lies strictly between $\lambda_i^{1/4}$ and $1$ whenever $\gamma>0$.\footnote{Appendix~\ref{app:two-layer-selection} proves uniqueness and these bounds.}
Thus the on-manifold prediction approaches, but does not reach, the zero-weight solution at any finite weight decay strength.

\subsection{Deeper chains confine anisotropy to the boundary layers}\label{sec:deep}

We now consider networks deeper than two layers, without weight decay, and ask how the noise-dependent structure is distributed across layers.
We will see that the entropic loss perspective predicts that the first and last layers have noise-dependent singular value profiles, while the interior layers have equal singular values within each layer.
We explain this structure first at depth $D=3$, where $\tW_3\tW_2\tW_1=I$.
See Appendix~\ref{app:deep} for a generalization to arbitrary depth, and Appendix~\ref{app:deep-gradient-balance} for the underlying Noether-style symmetry argument.

On the identity manifold,
$\tW_3\tW_2=\tW_1^{-1}$ and
$\tW_3=(\tW_2\tW_1)^{-1}$.
The three per-example gradients can therefore be written as
\begin{equation}\label{eq:deep-three-gradients}
    \nabla_{W_1}\ell=-2\,\tW_1^{-\top}\varepsilon x\T,
    \qquad
    \nabla_{W_2}\ell=-2\,(\tW_2\tW_1)^{-\top}\varepsilon(\tW_1x)\T,
    \qquad
    \nabla_{W_3}\ell=-2\,\varepsilon(\tW_2\tW_1x)\T.
\end{equation}
Using the same outer-product calculation as in
Section~\ref{sec:two-layer-selection}, their expected squared norms are
\begin{equation}\label{eq:deep-three-gradient-norms}
\begin{aligned}
    \E\|\nabla_{W_1}\ell\|_F^2
    &=4d\,\Tr\!\big(\tW_1^{-\top}\Sigeps\tW_1^{-1}\big),\\
    \E\|\nabla_{W_2}\ell\|_F^2
    &=4\,\Tr\!\big((\tW_2\tW_1)^{-\top}\Sigeps
        (\tW_2\tW_1)^{-1}\big)
        \Tr(\tW_1\T\tW_1),\\
    \E\|\nabla_{W_3}\ell\|_F^2
    &=4d\,\Tr\!\big((\tW_2\tW_1)\T(\tW_2\tW_1)\big).
\end{aligned}
\end{equation}

These expressions depend only on the Gram matrices of the cumulative transformations
through the first layer and through the first two layers. Define
\begin{equation}\label{eq:deep-three-gram-definitions}
    \Gamma_1:=\tW_1\T\tW_1,
    \qquad
    \Gamma_2:=(\tW_2\tW_1)\T(\tW_2\tW_1).
\end{equation}
Then, summing the three gradient norms and multiplying by $\eta/(4B)$ gives
\begin{equation}\label{eq:S-three}
    S(\Gamma_1,\Gamma_2)
    =\frac{\eta}{B}\left[
      d\Tr(\Sigeps\Gamma_1^{-1})
      +\Tr(\Sigeps\Gamma_2^{-1})\Tr\Gamma_1
      +d\Tr\Gamma_2\right].
\end{equation}

Setting the two matrix derivatives to zero gives
\begin{equation}\label{eq:deep-grams}
    \Gamma_1^2
    =\frac{d}{\Tr(\Sigeps\Gamma_2^{-1})}\Sigeps,
    \qquad
    \Gamma_2^2
    =\frac{\Tr\Gamma_1}{d}\Sigeps.
\end{equation}
Thus both cumulative Gram matrices are proportional to $\Sigeps^{1/2}$.
Solving for the constants yields
\begin{equation}\label{eq:deep-three-gram-minimum}
    \Gamma_1=c^{-1}\Sigeps^{1/2},
    \qquad
    \Gamma_2=c\Sigeps^{1/2},
    \qquad
    c:=\left(\frac{\Tr\Sigeps^{1/2}}{d}\right)^{1/3}.
\end{equation}
These Gram matrices attain the global minimum of Equation~\ref{eq:S-three}.

As in the two-layer case, these Gram matrices determine the cumulative transformations
up to orthogonal factors:
\begin{equation}\label{eq:deep-three-cumulative}
    \tW_1=c^{-1/2}U_1\Sigeps^{1/4},
    \qquad
    \tW_2\tW_1=c^{1/2}U_2\Sigeps^{1/4}.
\end{equation}
Both cumulative transformations contain the same factor $\Sigeps^{1/4}$.
Writing the middle layer as $\tW_2=(\tW_2\tW_1)\tW_1^{-1}$ cancels this factor, leaving $\tW_2=cU_2U_1\T$.

Together with $\tW_3=(\tW_2\tW_1)^{-1}$, this gives
\begin{equation}\label{eq:deep-three}
    \tW_1=c^{-1/2}U_1\Sigeps^{1/4},
    \qquad
    \tW_2=cU_2U_1\T,
    \qquad
    \tW_3=c^{-1/2}\Sigeps^{-1/4}U_2\T.
\end{equation}
Hence the first and last layers carry singular values proportional to
$\lambda_i^{1/4}$ and $\lambda_i^{-1/4}$, respectively, while every singular value
of the middle layer equals $c$.

\begin{figure}[!htbp]
    \centering
    \includegraphics[width=0.76\linewidth]{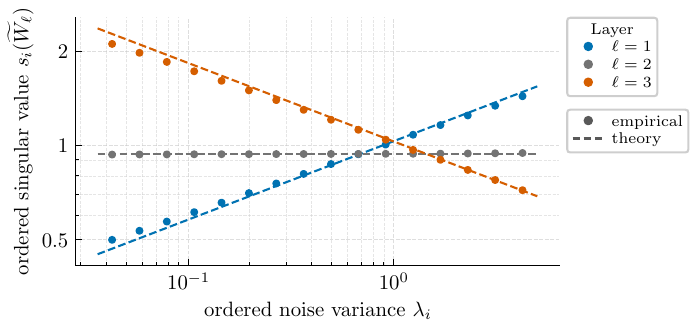}
    \caption{\textbf{In deeper networks, noise anisotropy is confined to the first and last layers.}
    Singular values of the three effective layers in a depth $D=3$ network, plotted
    against the noise variances.
    The first layer follows a fourth-root law, the middle layer has a flat spectrum,
    and the last layer follows an inverse fourth-root law.
    Points show mean singular values over six runs; error bars show one standard deviation.
    Dashed lines show the corresponding
    predictions $c^{-1/2}\lambda_i^{1/4}$, $c$, and $c^{-1/2}\lambda_i^{-1/4}$.
    Within each run, the first- and middle-layer singular values are sorted in ascending order, and the last-layer singular values in descending order, before pairing them with ascending noise variances.
    Here $d=16$, $\eta=0.02$, $B=64$, $\gamma=0$, and
    $1.5\times10^6$ SGD steps; weights are averaged over the final $1\%$ of training.}
    \label{fig:deep}
\end{figure}

\FloatBarrier
The same cancellation occurs at greater depths: the anisotropic dependence remains in
the first and last layers, while every interior layer is a scalar multiple of an
orthogonal matrix.
Figure~\ref{fig:deep} supports the $D=3$ prediction.

\subsection{Factoring the residual weights selects the zero-weight solution}
\label{sec:factored}

We return to the two-layer linear residual network without weight decay.
We now change only the parametrization by writing
\begin{equation}\label{eq:factored-parametrization}
    W_\ell=A_\ell B_\ell.
\end{equation}
The function class is unchanged, but the gradients with respect to the trainable parameters are now different.
Writing $G_\ell:=\nabla_{W_\ell}\ell$, the chain rule gives
\begin{equation}\label{eq:factored-chain-rule}
    \nabla_{A_\ell}\ell=G_\ell B_\ell\T,
    \qquad
    \nabla_{B_\ell}\ell=A_\ell\T G_\ell.
\end{equation}
Thus each factor appears in the gradient of the other.

At the zero-weight solution, $A_\ell=B_\ell=0$ for both layers, the \emph{residual maps} $h\mapsto A_\ell B_\ell h$ vanish and the skip connections implement the identity.
Every per-example gradient with respect to the factors also vanishes, so the entropic term is zero.
Conversely, for positive-definite input and noise covariances, the entropic term on the identity manifold vanishes only when $A_\ell=B_\ell=0$ for both layers -- any nonzero factor makes the expected squared gradient norm of the other factor in the same layer strictly positive (Appendix~\ref{app:factored-proof}).
Thus the zero-weight solution is the unique minimizer of the entropic term on the identity manifold, for either isotropic or anisotropic noise.

\begin{figure}[!htbp]
    \centering
    \includegraphics[width=0.76\linewidth]{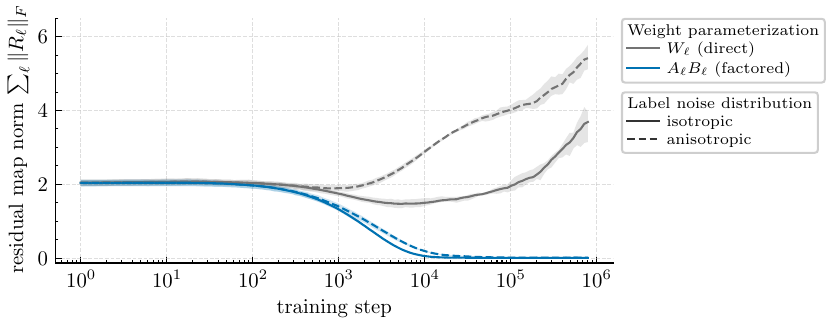}
    \caption{\textbf{Factoring the residual weights causes the residual branch to collapse to zero.}
    We compare direct residual weights $W_\ell$ with factored residual weights
    $W_\ell=A_\ell B_\ell$, starting from matched nonzero residual maps on the
    identity manifold.
    The y-axis shows the total residual map norm
    $\sum_\ell \|R_\ell\|_F$, where $R_\ell=W_\ell$ for the direct parametrization
    and $R_\ell=A_\ell B_\ell$ for the factored parametrization.
    Gray curves show direct weights and blue curves show factored weights;
    solid and dashed lines denote isotropic and anisotropic label noise, respectively.
    In both noise settings, the factored
    residual collapses toward zero while the direct residual remains nonzero.
    Lines show means over six runs and shaded regions show their ranges.
    Here $d=16$, $\eta=0.02$, $B=64$, and $\gamma=0$.}
    \label{fig:flip}
\end{figure}

\FloatBarrier
This predicts a different outcome from the direct parametrization, despite the function class being unchanged.
Experiments initialized with identical nonzero residual maps support this prediction: the factored residual maps collapse toward zero, while the directly parametrized residual maps remain nonzero (Figure~\ref{fig:flip}).
We also observe collapse of the residual branches in networks with ReLU blocks $h\mapsto h+A_\ell\phi(B_\ell h)$, where $\phi=\mathrm{ReLU}$ (Table~\ref{tab:factored} and Figure~\ref{fig:relu-functional}, Appendix~\ref{app:factored-experiments}).
These experiments extend the empirical observation beyond linear networks.

\section{Related work}\label{sec:related}

\paragraph{Modified losses and effective landscapes.}
Backward error analysis relates discrete optimization to a nearby continuous flow \citep{hairer2006geometric}. For deterministic full-batch gradient descent, the leading-order finite-step correction yields the modified objective $\mathcal L+\frac{\eta}{4}\|\nabla\mathcal L\|^2$ \citep{barrett2021implicit}. \citet{smith2021origin} derive a related minibatch-gradient correction for SGD with random reshuffling over a finite dataset. \citet{ziyin2025neural} apply this construction to a fixed minibatch and call the batch-averaged effective objective the \emph{entropic loss}.  We adopt this viewpoint for our online setting and test its predictions. Classical analyses of stochastic approximation \citep{robbins1951stochastic,bottou1998online,bottou2018optimization} characterize SGD's convergence and the noise-induced fluctuation around a minimum. Complementary approaches analyze stochastic dynamics and stationary distributions directly \citep{li2019stochastic,mandt2017stochastic,yaida2019fluctuation,liu2021noise}, including motion along manifolds of minimizers \citep{li2022happens}. \citet{blanc2020implicit} characterize the implicit regularization induced by label noise near zero-training-error solutions.

Sharpness-aware minimization (SAM) explicitly modifies the optimization objective to favor solutions whose loss remains low under small parameter perturbations \citep{foret2021sharpness}. Its first-order approximation yields a penalty proportional to the minibatch-gradient norm, closely related to the squared-norm penalty in the entropic objective. Our use of the entropic objective serves a different purpose: we use it to predict the implicit bias of ordinary SGD, without explicitly adding the penalty during training.

\paragraph{Symmetry, noise equilibrium, and entropic selection.}
Parameter symmetries yield conservation laws and initialization-dependent
layer balance under deterministic gradient flow
\citep{du2018algorithmic,kunin2021neural}.
Under stochastic gradients, the
same symmetry directions acquire systematic noise-induced motion.
\citet{ziyin2024parameter} describe this motion as a Noether flow and define
noise equilibria by a balance of gradient noise across symmetry-related
parameter directions. They derive the resulting alignment conditions for
matrix factorizations and exact noise-equilibrium solutions for deep linear
networks. Our two-layer identity manifold is the orbit of exactly such a symmetry;
Appendix~\ref{app:deep-gradient-balance} derives the equilibrium condition.
We revisit a different instance of this rescaling symmetry in
Section~\ref{sec:factored}, where the entropic-term argument instead
predicts collapse to the zero-weight solution.
Relatedly, this zero-weight solution is an invariant point of SGD in the sense of \citet{chen2023stochastic}, who study how gradient noise can attract training toward invariant sets (``stochastic collapse'').
\citet{ziyin2025neural} formulate an entropic loss landscape that
breaks continuous parameter symmetries, derive layer- and neuron-level
gradient-balance relations, and use them to study universal representation
alignment and sharpening or flattening.
More recently, \citet{aladrah2026understanding} give a complementary geometric
account of stochastic implicit bias and its control through
predictor-preserving reparameterizations.

\paragraph{Linear models and factorization.}
Deep linear networks have long provided exactly solvable models of learning
dynamics \citep{saxe2013exact,rhee-2026-deep-linear-nets}.
\citet{pmlr-v202-pashakhanloo23a} study a closely related setting: continual
online SGD in a two-layer linear autoencoder with an expansive hidden layer,
trained on the clean identity target $y=x$. They include $L_2$ regularization; the remaining minimum-loss
solutions are rotationally equivalent hidden representations, and randomness
from sampling one input at a time produces normal fluctuations and tangential
diffusion along this manifold. Their analysis focuses on the rate and stimulus
dependence of this continuing representational drift, which vanishes in their
model when the regularization coefficient is zero. We instead add irreducible
label noise, $y=x+\varepsilon$, so stochastic gradients remain nonzero at the
exact identity solution even without weight decay, and ask how they select
among factorizations with different scales and singular spectra.
This connection also suggests potential relevance to representational drift in biological systems, including olfaction, although we do not investigate that application here.
\citet{xu2026doessgdseekflatness}
analyze the sharpness selected under the minimum-gradient-fluctuation
condition: isotropic label noise selects the minimum-sharpness solution,
whereas anisotropic label noise can select a sharper one.

\paragraph{Residual networks.}
Following earlier work on gated shortcuts in highway networks \citep{srivastava2015training}, \citet{he2016deep} introduced deep residual networks.
As motivation, they present the puzzling phenomenon of deeper networks resulting in higher training error than shallower ones; in principle, a deeper network could simply match a shallower network by adding layers that implement the identity, suggesting that the difficulty lay in finding such a solution through optimization.
They therefore proposed parametrizing blocks as $x+F(x)$, hypothesizing that driving $F(x)$ to zero would be easier than learning the identity directly.
Residual connections have since become a cornerstone of deep learning architectures, including deep convolutional networks for image recognition \citep{szegedy2017inception} and transformer language models \citep{vaswani2017attention,brown2020language}.
\citet{veit2016residual} interpret residual networks as collections of paths and provide empirical evidence of behavior resembling ensembles of relatively shallow networks.
The residual stream is also a central object of study in mechanistic interpretability: \citet{elhage2021mathematical} describe it as a communication channel through which attention heads and MLPs read and write information.

\section{Discussion}\label{sec:discussion}

Broadly, our results illustrate how gradient noise can influence the internal decomposition of a learned function.
In general, a deep network can be decomposed into a composition of two (and usually more than two) functions:
\begin{equation}
    f= g_2 \circ  g_1,
\end{equation}
where $g_1(x)$ is a latent representation and $g_2$ represents the downstream layers, mapping the representation to the output. In principle, for any invertible $h$, the same input--output function admits the alternative decomposition
\begin{equation}
    g_1\to h\circ g_1,\qquad g_2\to g_2\circ h^{-1}.
\end{equation}
The input--output function therefore does not uniquely determine the learned representation; which decomposition is learned depends on initialization and training.
In this paper, we showed that, in the settings studied here, networks can learn a decomposition of $f$ that differs from the trivial layerwise implementation: even when learning an identity function, they compose two functions that need not individually be close to the identity.
Under anisotropic label noise, this effect, perhaps surprising to many, persists even if we introduce a regularization term that favors the zero-weight implementation of the identity.

\paragraph{Limitations.}
In this manuscript, we focus on small, analytically solvable toy models.
Within these simple models, we showed that SGD favors characteristic solution structures that are robust across different initializations, with individual layers often far away from the identity.
We demonstrated some qualitatively similar effects on nonlinear models (Appendix~\ref{app:factored-experiments}), but it remains an interesting future problem to extend the theory to nonlinear problems.
Our experiments use online SGD with fresh samples and a constant learning rate; extending the analysis to finite datasets, momentum, adaptive optimizers, or learning-rate schedules is also an open problem. Our entropic loss analysis includes only the lowest-order correction in the neural thermodynamics framework; understanding how higher-order terms affect its predictions remains an open problem.

Our theoretical analysis characterizes minimizers of the entropic objective but does not establish convergence of SGD to these minimizers; the correspondence is supported empirically in the principal settings studied here. In general, we do not yet understand when minimizing the entropic loss accurately predicts the solutions favored by SGD.
Our experiments with noncommuting covariances show discrepancies between the predicted and learned Gram matrices (Appendix~\ref{app:noncommuting}).
Establishing the regime of validity of entropic loss minimization as a predictor of SGD solutions is an important direction for future work.

\section*{Acknowledgments}
AA and DB are supported by grants from Coefficient Giving.
DB is supported by NSF \#2408455.
LZ thanks the generous support from NTT Research.

\bibliographystyle{tmlr}
\bibliography{ref}

\newpage
\appendix

\etocdepthtag.toc{appendices}
\begingroup
\hypersetup{linkcolor=black}
\etocsettagdepth{body}{none}
\etocsettagdepth{appendices}{subsection}
\etocsetnexttocdepth{subsection}
\etocsettocstyle{\section*{Appendix}\small\setlength{\parskip}{0pt}}{}
\tableofcontents
\endgroup
\clearpage

\section{Population loss and gradient calculations}\label{app:loss-deriv}
We consider the two-layer network $f(x)=\tW_2\tW_1x$, with $\tW_\ell=I+W_\ell$, and labels $y=x+\varepsilon$.
Inputs and label noise are independent and mean zero, with covariances $\Sigx\succ0$ and $\Sigeps\succ0$.

\subsection{Population loss and the identity manifold}\label{app:population-loss}
Let $M:=\tW_2\tW_1-I$. The prediction error is $y-f(x)=\varepsilon-Mx$,
so the population loss is
\begin{align*}
    \mathcal L
    &=\E\|\varepsilon-Mx\|^2\\
    &=\E\|\varepsilon\|^2-2\E[\varepsilon\T Mx]+\E[x\T M\T Mx]\\
    &=\Tr\Sigeps+\Tr(M\Sigx M\T).
\end{align*}
The cross term vanishes by independence and zero means. The other two terms
follow from $\E[z\T Qz]=\Tr(Q\Sigma_z)$ for a mean-zero vector $z$ with
covariance $\Sigma_z$.

Since $\Sigx\succ0$, the term
$\Tr(M\Sigx M\T)=\|M\Sigx^{1/2}\|_F^2$ vanishes exactly when $M=0$.
Thus the minimum population loss is $\Tr\Sigeps$, attained on
\[
    \M=\{(W_1,W_2):\tW_2\tW_1=I\}.
\]

For $\Sigx=I$, substituting into the population loss gives
\begin{align*}
    \mathcal L
    &=\Tr\Sigeps+\Tr(MM\T)\\
    &=\Tr\Sigeps+\|M\|_F^2\\
    &=\Tr\Sigeps+\|\tW_2\tW_1-I\|_F^2.
\end{align*}

\subsection{Per-example and population gradients}\label{app:population-gradients}
Write $h:=\tW_1x$ and $r:=y-\tW_2h$, so the per-example loss is
$\ell=\|r\|^2$. Differentiating the second layer gives
$\nabla_{W_2}\ell=-2rh\T$. For the first layer,
$\partial\ell/\partial h=-2\tW_2\T r$, so
\[
    \nabla_{W_1}\ell=-2(\tW_2\T r)x\T,
    \qquad
    \nabla_{W_2}\ell=-2r(\tW_1x)\T.
\]
Derivatives with respect to $W_\ell$ and $\tW_\ell$ coincide because
$\tW_\ell=I+W_\ell$.

Using $r=\varepsilon-Mx$ and $\E[rx\T]=-M\Sigx$, the population
gradients are
\[
    \nabla_{W_1}\mathcal L=2\tW_2\T M\Sigx,
    \qquad
    \nabla_{W_2}\mathcal L=2M\Sigx\tW_1\T.
\]
For $\Sigx=I$, these reduce to $2\tW_2\T M$ and $2M\tW_1\T$.
Both population gradients vanish on $\M$. The per-example gradients generally
remain nonzero because the prediction error still contains label noise, $r=\varepsilon$.

\subsection{Expected squared gradient norms on the identity manifold}\label{app:gradient-norms}
On $\M$, we have $r=\varepsilon$. Substituting into the gradients from
Appendix~\ref{app:population-gradients}, using
$\|uv\T\|_F^2=\|u\|^2\|v\|^2$ and independence of $x$ and $\varepsilon$, gives
\begin{align*}
    \E\|\nabla_{W_1}\ell\|_F^2
    &=4\E\|\tW_2\T\varepsilon\|^2\,\E\|x\|^2
      =4\Tr(\tW_2\T\Sigeps\tW_2)\Tr\Sigx,\\
    \E\|\nabla_{W_2}\ell\|_F^2
    &=4\E\|\varepsilon\|^2\,\E\|\tW_1x\|^2
      =4\Tr\Sigeps\,\Tr(\tW_1\Sigx\tW_1\T).
\end{align*}
The trace expressions follow from $\E\|Qz\|^2=\Tr(Q\Sigma_zQ\T)$.
With $\Sigx=I$ and $\Tr\Sigeps=d$, summing over the two layers yields
\[
    \E\|\nabla\ell\|^2
    =4d\left[\Tr(\tW_1\tW_1\T)
      +\Tr(\tW_1^{-\top}\Sigeps\tW_1^{-1})\right],
\]
where we substituted $\tW_2=\tW_1^{-1}$. Multiplying by $\eta/(4B)$,
as in Equation~\ref{eq:S-on-M}, gives the entropic term in
Equation~\ref{eq:S}.

\section{Derivation of the entropic loss}\label{app:entropic}
\subsection{Correction for a finite step size}\label{app:finite-step}
Let $J(\theta)$ be a smooth loss, and write $g:=\nabla J(\theta)$ and
$H:=\nabla^2J(\theta)$. Gradient flow starting at $\theta$ has initial
derivatives $\dot\theta(0)=-g$ and $\ddot\theta(0)=Hg$, so after time $\eta$,
\[
    \theta(\eta)=\theta-\eta g+\frac{\eta^2}{2}Hg+O(\eta^3).
\]
One gradient descent step gives $\theta-\eta g$. To cancel the extra
term of order $\eta^2$, consider gradient flow on $J+\eta R$. Its expansion is
\[
    \theta(\eta)=\theta-\eta g-\eta^2\nabla R(\theta)
    +\frac{\eta^2}{2}Hg+O(\eta^3).
\]
The two updates agree through order $\eta^2$ when
\[
    \nabla R=\frac12Hg
    =\nabla\left(\frac14\|\nabla J\|^2\right).
\]
Thus gradient flow on the modified loss
\[
    J_\eta:=J+\frac\eta4\|\nabla J\|^2
\]
matches one gradient descent step on $J$ up to $O(\eta^3)$ over time $\eta$.
Backward error analysis describes a discrete numerical method through a
modified differential equation; see \citet{hairer2006geometric} for the
general framework. \citet{barrett2021implicit} apply this approach to
gradient descent and derive the gradient norm correction above.

\subsection{Minibatch averaging and the entropic term}\label{app:minibatch-averaging}
For a fixed minibatch $\mathcal B$ of $B$ independent examples, apply the
calculation above to $J=\ell_{\mathcal B}:=\frac{1}{B}\sum_{b=1}^B\ell_b$.
Averaging this modified loss over minibatches yields
the \emph{entropic loss} \citep{ziyin2025neural}:
\[
    F_\eta:=\mathcal L+S,
    \qquad
    S:=\frac\eta4\E_{\mathcal B}\|\nabla\ell_{\mathcal B}\|^2.
\]

At fixed parameters, the squared minibatch gradient is
\begin{align*}
    \|\nabla\ell_{\mathcal B}\|^2
    &=\left\|\frac1B\sum_{b=1}^B\nabla\ell_b\right\|^2\\
    &=\frac1{B^2}
      \left(\sum_{b=1}^B\nabla\ell_b\right)\T
      \left(\sum_{b'=1}^B\nabla\ell_{b'}\right)\\
    &=\frac1{B^2}\sum_{b=1}^B\sum_{b'=1}^B
      (\nabla\ell_b)\T\nabla\ell_{b'}\\
    &=\frac1{B^2}\left[
      \sum_{b=1}^B\|\nabla\ell_b\|^2
      +\sum_{b\ne b'}(\nabla\ell_b)\T\nabla\ell_{b'}\right].
\end{align*}
For $b\ne b'$, independence gives
\[
    \E[(\nabla\ell_b)\T\nabla\ell_{b'}]
    =(\E\nabla\ell_b)\T(\E\nabla\ell_{b'})
    =(\nabla\mathcal L)\T\nabla\mathcal L
    =\|\nabla\mathcal L\|^2.
\]

Taking expectations in the expansion above yields
\begin{align*}
    \E_{\mathcal B}\|\nabla\ell_{\mathcal B}\|^2
    &=\frac1{B^2}\left[B\E\|\nabla\ell\|^2
      +B(B-1)\|\nabla\mathcal L\|^2\right]\\
    &=\left(1-\frac1B\right)\|\nabla\mathcal L\|^2
      +\frac1B\E\|\nabla\ell\|^2.
\end{align*}
On the identity manifold,
$\nabla\mathcal L=0$, so this reduces to
\[
    S=\frac{\eta}{4B}\E\|\nabla\ell\|^2,
\]
as used in Equation~\ref{eq:S-on-M}.

\subsection{Including weight decay}\label{app:entropic-weight-decay}
Adding $\gamma\|\theta\|^2$ to each minibatch loss changes its gradient to
$\nabla\ell_{\mathcal B}+2\gamma\theta$. To first order in $\eta$, the
entropic loss is therefore
\begin{align*}
    F_{\eta,\gamma}
    &:=\mathcal L+\gamma\|\theta\|^2
      +\frac\eta4\E_{\mathcal B}
        \|\nabla\ell_{\mathcal B}+2\gamma\theta\|^2\\
    &=\mathcal L+S+\gamma(1+\eta\gamma)\|\theta\|^2
      +\eta\gamma\,\theta\T\nabla\mathcal L.
\end{align*}
On $\M$, the last term vanishes, giving Equation~\ref{eq:Lent-wd-M}:
\[
    F_{\eta,\gamma}
    =\mathcal L+S+\gamma(1+\eta\gamma)\|\theta\|^2.
\]
Here $\theta$ consists of the residual weights, so
$\|\theta\|^2=\|W_1\|_F^2+\|W_2\|_F^2$.
The relative correction to the weight decay coefficient is $\eta\gamma$,
which is at most $0.02$ in our experiments. The body generally neglects
this correction; although the quartic predictions plotted in Figure~\ref{fig:fourthroot} retain it.

\subsection{The entropic loss away from the identity manifold}
\label{app:offman}
Assume $x\sim\mathcal N(0,I)$ and label noise independent of $x$, with
mean zero and covariance $\Sigeps$. Using $M=\tW_2\tW_1-I$ and
$r=\varepsilon-Mx$ in the gradients from
Appendix~\ref{app:population-gradients} gives
\begin{align*}
    \E\|\nabla_{W_1}\ell\|_F^2
    &=4d\Tr(\tW_2\T\Sigeps\tW_2)
      +4\E\big[\|\tW_2\T Mx\|^2\|x\|^2\big],\\
    \E\|\nabla_{W_2}\ell\|_F^2
    &=4\Tr\Sigeps\,\|\tW_1\|_F^2
      +4\E\big[\|Mx\|^2\|\tW_1x\|^2\big].
\end{align*}
The cross terms vanish because $\varepsilon$ is independent of $x$ and
mean zero. For symmetric matrices $Q$ and $R$, the Gaussian fourth moment
identity is
\[
    \E[(x\T Qx)(x\T Rx)]=\Tr Q\Tr R+2\Tr(QR).
\]
Applying it with $(Q,R)=(M\T\tW_2\tW_2\T M,I)$ for the first layer
and $(Q,R)=(M\T M,\tW_1\T\tW_1)$ for the second yields
\begin{align*}
    \E\|\nabla_{W_1}\ell\|_F^2
    &=4\left[d\Tr(\tW_2\T\Sigeps\tW_2)
      +(d+2)\|\tW_2\T M\|_F^2\right],\\
    \E\|\nabla_{W_2}\ell\|_F^2
    &=4\left[(\Tr\Sigeps+\|M\|_F^2)\|\tW_1\|_F^2
      +2\Tr(M\T M\tW_1\T\tW_1)\right].
\end{align*}
Together with
\[
    \|\nabla\mathcal L\|^2
    =4\|\tW_2\T M\|_F^2+4\|M\tW_1\T\|_F^2,
\]
these give the entropic term away from $\M$ through
\[
    S=\frac\eta4\left[
      \left(1-\frac1B\right)\|\nabla\mathcal L\|^2
      +\frac1B\E\|\nabla\ell\|^2\right].
\]
Setting $M=0$ recovers Equation~\ref{eq:S} under $\Tr\Sigeps=d$.

\section{Two-layer solution selection and weight decay}\label{app:two-layer-selection}
Throughout this appendix, $\Sigeps\succ0$, $\Tr\Sigeps=d$, and
$\kappa:=\eta d/B>0$.

\subsection{Strict convexity of the Gram objective}\label{app:gram-derivative}
Section~\ref{sec:two-layer-selection} derives the objective in
Equation~\ref{eq:S-gram} and its stationary point $G=\Sigeps^{1/2}$.
To establish strict convexity, take any $G\succ0$ and nonzero symmetric
perturbation $\Delta$. Using
$\frac{d}{dt}(G+t\Delta)^{-1}
=-(G+t\Delta)^{-1}\Delta(G+t\Delta)^{-1}$ gives
\begin{align*}
    \frac{d^2}{dt^2}\Big|_{t=0}S(G+t\Delta)
    &=2\kappa\Tr(\Sigeps G^{-1}\Delta G^{-1}\Delta G^{-1})\\
    &=2\kappa\big\|\Sigeps^{1/2}G^{-1}\Delta G^{-1/2}\big\|_F^2>0.
\end{align*}
The inequality is strict because $\Sigeps$ and $G$ are invertible.
Thus $S$ is strictly convex on the convex domain $G\succ0$, and the
stationary point found in the body is its unique global minimizer.

\subsection{Diagonal reduction with weight decay}\label{app:wd-diagonal}
On $\M$, write the polar decomposition $\tW_1=OP$, where $O$ is
orthogonal and $P\succ0$. Then $\tW_2=P^{-1}O\T$.
The Gram matrix is $P^2$, so Equation~\ref{eq:S-gram} is independent of $O$.
The residual weight penalty is
\begin{align*}
    \|W_1\|_F^2+\|W_2\|_F^2
    &=\|OP-I\|_F^2+\|P^{-1}O\T-I\|_F^2\\
    &=\Tr(P^2+P^{-2})+2d-2\Tr\big[O(P+P^{-1})\big].
\end{align*}
In an eigenbasis of $P$, with eigenvalues $a_i>0$,
\[
    \Tr\big[O(P+P^{-1})\big]
    =\sum_i O_{ii}(a_i+a_i^{-1})
    \le\sum_i(a_i+a_i^{-1}),
\]
since $O_{ii}\le1$. Equality requires $O=I$. Thus, for $\gamma>0$,
minimizing over $O$ makes $\tW_1=P$ symmetric positive definite.

For fixed eigenvalues of $P$, the penalty and $\Tr P^2$ are constant;
only $\Tr(\Sigeps P^{-2})$ depends on its eigenvectors. Order the noise
eigenvalues and the $a_i$ so that
$\lambda_1\ge\cdots\ge\lambda_d$ and $a_1\ge\cdots\ge a_d$.
The trace rearrangement inequality gives
\[
    \Tr(\Sigeps P^{-2})\ge\sum_i\lambda_i a_i^{-2},
\]
with equality when $P$ is diagonal in the noise eigenbasis, pairing
$\lambda_i$ with $a_i$. A global minimizer can therefore be chosen
diagonal and positive, as assumed in Equation~\ref{eq:phi}.

\subsection{Uniqueness and location of the positive quartic root}
\label{app:quartic-numerics}
For the scalar objective $\phi_i$ in Equation~\ref{eq:phi},
\[
    \phi_i''(a)
    =2\kappa(1+3\lambda_i a^{-4})
     +2\gamma(1+3a^{-4}-2a^{-3})>0
    \qquad(a>0).
\]
Indeed, multiplying the second parenthesis by $a^4$ gives
\[
    a^4+3-2a=(a-1)^2(a^2+2a+3)+2a>0.
\]
Since $\phi_i$ also diverges as $a\to0$ and $a\to\infty$, it has a
unique minimizer. Equation~\ref{eq:quartic} is equivalent to
$\phi_i'(a)=0$ for $a>0$, so it has exactly one positive root.

To locate this root, set $q_i:=\lambda_i^{1/4}$. At the two values
favored separately by weight decay and the entropic term,
\[
    \phi_i'(1)=2\kappa(1-\lambda_i),
    \qquad
    \phi_i'(q_i)=2\gamma(q_i-1)(1+q_i^{-3}).
\]
For $\gamma>0$ and $\lambda_i\ne1$, these have opposite signs.
The unique root therefore lies strictly between $1$ and $q_i$.
If $\lambda_i=1$, the root is $1$ for every $\gamma\ge0$;
if $\gamma=0$, it is $q_i$.

For Figure~\ref{fig:fourthroot}, we compute the roots of
Equation~\ref{eq:quartic} using the experimental hyperparameters and
retain the positive real root. The plotted predictions replace $\gamma$
by $\gamma(1+\eta\gamma)$, retaining the correction derived in
Appendix~\ref{app:entropic-weight-decay}. This positive rescaling leaves the arguments
above unchanged.

\section{Deeper networks}\label{app:deep}
We extend the calculation in Section~\ref{sec:deep} to depth $D\ge3$,
with $\Sigx=I$, $\Sigeps\succ0$, $\Tr\Sigeps=d$, and no weight decay.

\subsection{Cumulative Gram matrices and the entropic term}\label{app:deep-objective}
Write
\[
    C_k:=\tW_k\cdots\tW_1,
    \qquad \Gamma_k:=C_k\T C_k,
    \qquad C_0=C_D=I.
\]
On the identity manifold, the product of the layers after layer $\ell$ is
$\tW_D\cdots\tW_{\ell+1}=C_\ell^{-1}$. The gradient for one example is
therefore
\[
    \nabla_{W_\ell}\ell
    =-2C_\ell^{-\top}\varepsilon(C_{\ell-1}x)\T.
\]
Using the expectation calculation in Appendix~\ref{app:gradient-norms},
\[
    \E\|\nabla_{W_\ell}\ell\|_F^2
    =4\Tr(\Sigeps\Gamma_\ell^{-1})\Tr\Gamma_{\ell-1}.
\]
Thus Equation~\ref{eq:S-on-M} gives
\begin{equation}\label{eq:S-deep}
    S=\frac\eta B\sum_{\ell=1}^D
      \Tr(\Sigeps\Gamma_\ell^{-1})\Tr\Gamma_{\ell-1},
    \qquad \Gamma_0=\Gamma_D=I.
\end{equation}
Every choice of $\Gamma_1,\dots,\Gamma_{D-1}\succ0$ is feasible:
take $C_k=\Gamma_k^{1/2}$ and $\tW_k=C_kC_{k-1}^{-1}$.
Thus minimizing over the layers is equivalent to minimizing over
$\Gamma_1,\dots,\Gamma_{D-1}\succ0$.

\subsection{Global minimizers and layer structure}\label{app:deep-minimum}
Set $\tau:=\Tr\Sigeps^{1/2}$ and $t_k:=\Tr\Gamma_k$.
For each $k=1,\dots,D-1$, matrix Cauchy--Schwarz gives
\begin{align*}
    \tau^2
    &=\left[\Tr\big(\Gamma_k^{1/2}
        \Gamma_k^{-1/2}\Sigeps^{1/2}\big)\right]^2\\
    &\le\|\Gamma_k^{1/2}\|_F^2
          \|\Gamma_k^{-1/2}\Sigeps^{1/2}\|_F^2\\
    &=t_k\Tr(\Sigeps\Gamma_k^{-1}).
\end{align*}
Equality holds exactly when $\Gamma_k$ is proportional to
$\Sigeps^{1/2}$. Substituting this bound into Equation~\ref{eq:S-deep}
leaves an expression involving only the traces:
\[
    \frac B\eta S
    \ge\frac{d\tau^2}{t_1}
       +\sum_{k=2}^{D-1}\frac{\tau^2t_{k-1}}{t_k}
       +dt_{D-1}.
\]
The product of these $D$ positive terms is independent of the $t_k$:
\[
    \frac{d\tau^2}{t_1}
    \left(\prod_{k=2}^{D-1}\frac{\tau^2t_{k-1}}{t_k}\right)
    dt_{D-1}=d^2\tau^{2(D-1)}.
\]
The arithmetic--geometric mean inequality therefore gives
\[
    S\ge\frac\eta B\,D d^{2/D}\tau^{2(D-1)/D}.
\]
Equality requires all $D$ terms to be equal. Writing
$c_D:=(\tau/d)^{1/D}$, their common value must be $\tau^2/c_D^2$.
The first term and then each successive term give
\[
    t_1=dc_D^2,
    \qquad t_k=c_D^2t_{k-1}=dc_D^{2k}
    \quad(2\le k\le D-1).
\]
Combining this with equality in Cauchy--Schwarz yields
\begin{equation}\label{eq:deep-minimizing-grams}
    \Gamma_k=\frac{t_k}{\tau}\Sigeps^{1/2}
      =c_D^{2k-D}\Sigeps^{1/2},
    \qquad 1\le k\le D-1.
\end{equation}
These matrices attain both bounds, so they are the unique minimizing
cumulative Gram matrices. For $D=3$, they are precisely the two matrices
in Equation~\ref{eq:deep-three-gram-minimum}.

To recover the individual layers, write
$C_k=U_k\Gamma_k^{1/2}
=c_D^{k-D/2}U_k\Sigeps^{1/4}$ for $1\le k\le D-1$,
with arbitrary orthogonal $U_k$ and $C_0=C_D=I$.
Taking $\tW_k=C_kC_{k-1}^{-1}$ gives all minimizing factorizations:
\begin{equation}\label{eq:deep-all-layers}
    \tW_\ell=\begin{cases}
        a_D U_1\Sigeps^{1/4}, & \ell=1,\\
        c_D U_\ell U_{\ell-1}\T, & 2\le\ell\le D-1,\\
        a_D\Sigeps^{-1/4}U_{D-1}\T, & \ell=D,
    \end{cases}
    \qquad
    a_D:=c_D^{-(D-2)/2}.
\end{equation}
The common covariance factor cancels between consecutive cumulative maps,
leaving every interior layer equal to $c_D$ times an orthogonal matrix.
It remains in the first and last layers, whose singular values are
$a_D\lambda_i^{1/4}$ and $a_D\lambda_i^{-1/4}$.

\subsection{Connection to gradient balance}\label{app:deep-gradient-balance}

\citet{ziyin2025neural} show that when two parameter matrices enter the loss
through their product, minimizing the leading-order entropic objective imposes
a balance between their gradients. Without weight decay, their gradient-alignment
result gives
\[
    \E[G_\ell G_\ell\T]=\E[G_{\ell+1}\T G_{\ell+1}],
    \qquad G_\ell:=\nabla_{W_\ell}\ell_{\mathcal B}.
\]
Taking traces shows that adjacent layers have equal expected squared gradient norms.

To understand this condition, consider rescaling two adjacent effective layers:
\[
    \tW_\ell\mapsto c\tW_\ell,
    \qquad \tW_{\ell+1}\mapsto c^{-1}\tW_{\ell+1},
    \qquad c>0.
\]
Their product, and hence the network's output, remains unchanged.
Their gradients instead transform as $G_\ell\mapsto c^{-1}G_\ell$ and
$G_{\ell+1}\mapsto cG_{\ell+1}$. All other gradients remain unchanged,
so the part of the entropic term affected by this rescaling is
\[
    \frac\eta4\left(
      c^{-2}\E\|G_\ell\|_F^2+c^2\E\|G_{\ell+1}\|_F^2
    \right).
\]
Here $c=1$ corresponds to the original weights. If these weights minimize the
entropic term, no small rescaling can reduce it. The derivative with respect
to $c$ must therefore vanish at $c=1$, giving equal expected squared gradient
norms.

To obtain the stronger matrix balance condition, we stretch an arbitrary
unit direction $u$ of the hidden representation and undo the stretch in the
next layer. Repeating the rescaling argument gives
\[
    \E\|G_\ell\T u\|_2^2=\E\|G_{\ell+1}u\|_2^2,
\]
or equivalently,
\[
    u\T\E[G_\ell G_\ell\T]u
    =u\T\E[G_{\ell+1}\T G_{\ell+1}]u.
\]
Since this equality holds for every direction $u$, the two symmetric matrices
must be equal. These transformations preserve the identity manifold, so the
balance condition also holds at local minima of the
entropic term restricted to this manifold.

We now apply this result using the cumulative maps
$C_k=\tW_k\cdots\tW_1$ and Gram matrices $\Gamma_k=C_k\T C_k$
from Appendix~\ref{app:deep-objective}. On the identity manifold, the gradient
formula there gives
\[
\begin{aligned}
    \E[G_\ell G_\ell\T]
    &=\frac4B\Tr(\Gamma_{\ell-1})
      C_\ell^{-\top}\Sigeps C_\ell^{-1},\\
    \E[G_{\ell+1}\T G_{\ell+1}]
    &=\frac4B\Tr(\Sigeps\Gamma_{\ell+1}^{-1})C_\ell C_\ell\T.
\end{aligned}
\]
Equating these expressions and multiplying on the left by $C_\ell\T$ and
on the right by $C_\ell$ yields
\[
    \Tr(\Gamma_{\ell-1})\Sigeps
    =\Tr(\Sigeps\Gamma_{\ell+1}^{-1})\Gamma_\ell^2,
    \qquad \Gamma_0=\Gamma_D=I.
\]
Thus every intermediate cumulative Gram matrix is proportional to
$\Sigeps^{1/2}$.

For $D=3$, this becomes
\[
    \Gamma_1^2=\frac{d}{\Tr(\Sigeps\Gamma_2^{-1})}\Sigeps,
    \qquad
    \Gamma_2^2=\frac{\Tr\Gamma_1}{d}\Sigeps,
\]
recovering the equations derived by direct minimization in
Section~\ref{sec:deep}. More generally, the common noise-dependent factor
cancels between consecutive cumulative maps, leaving the interior layers
proportional to orthogonal matrices. Gradient balance therefore recovers
the structure characterized in Appendix~\ref{app:deep-minimum}, where we
also establish global optimality.

\subsection{Experiments at greater depth}\label{app:deep-experiments}
\begin{figure}[ht]
    \centering
    \includegraphics[width=0.98\linewidth]{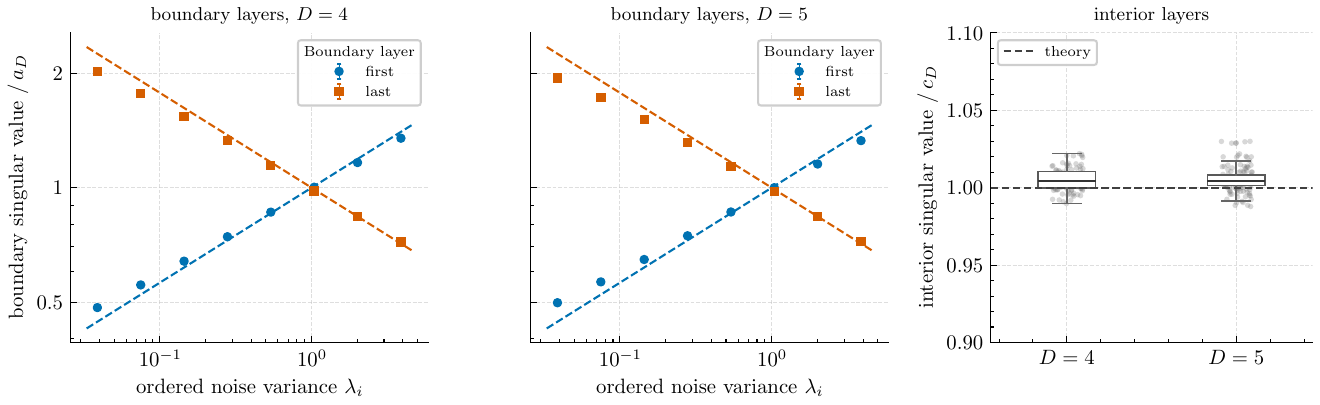}
    \caption{\textbf{Anisotropy remains confined to the boundary layers at greater depths.}
    \emph{Left and middle:} Mean boundary singular values for $D=4$ and $D=5$,
    respectively, divided by $a_D$ and plotted against the noise variance.
    Error bars show one standard deviation across runs. Dashed lines show
    $\lambda_i^{1/4}$ and $\lambda_i^{-1/4}$. Within each run, the first
    layer's singular values are sorted in ascending order and the last
    layer's in descending order, then paired with ascending noise variances.
    \emph{Right:} Interior singular values divided by $c_D$; the dashed line
    marks one. Points show individual singular values across layers, noise
    directions, and runs; boxes show the median and interquartile range.
    Each depth uses six runs with $d=8$, $\eta=0.02$, $B=64$, $\gamma=0$,
    and $1.5\times10^6$ SGD steps; singular values are computed from weights
    averaged over the final $1\%$ of training.}
    \label{fig:deep-extra}
\end{figure}

Figure~\ref{fig:deep-extra} tests Equation~\ref{eq:deep-all-layers} at
$D=4$ and $D=5$. Dividing the boundary singular values by $a_D$ leaves
the predictions $\lambda_i^{1/4}$ and $\lambda_i^{-1/4}$.
Dividing the interior singular values by $c_D$ gives a prediction of one
for every layer and noise direction.

\FloatBarrier
\section{Factored residual parametrizations}\label{app:factored}

\subsection{Uniqueness of the zero-factor minimizer}\label{app:factored-proof}
We use the two-layer linear network of Section~\ref{sec:factored}, with
$\tW_\ell=I+A_\ell B_\ell$ and $A_\ell,B_\ell\in\R^{d\times d}$.

\begin{proposition}\label{prop:collapse}
For independent, mean-zero inputs and label noise with
$\Sigx,\Sigeps\succ0$, and $\eta>0$, the entropic term on the identity
manifold is zero if and only if $A_1=B_1=A_2=B_2=0$.
This is its unique global minimizer on the identity manifold.
\end{proposition}

\begin{proof}
Apply the factor chain rule in Equation~\ref{eq:factored-chain-rule} to
the gradients in Appendix~\ref{app:population-gradients}, with
$r=\varepsilon$ on $\M$. The expectation calculation in
Appendix~\ref{app:gradient-norms} and the minibatch factor $\eta/(4B)$
from Equation~\ref{eq:S-on-M} give
\begin{align*}
    S=\frac{\eta}{B}\big[\,
    &\Tr(\tW_2\T\Sigeps\tW_2)\Tr(B_1\Sigx B_1\T)
    +\Tr(\Sigx)\Tr(A_1\T\tW_2\T\Sigeps\tW_2A_1)\\
    &+\Tr(\Sigeps)\Tr(B_2\tW_1\Sigx\tW_1\T B_2\T)
    +\Tr(\tW_1\Sigx\tW_1\T)\Tr(A_2\T\Sigeps A_2)\,\big].
\end{align*}
The four terms correspond to the gradients with respect to
$A_1,B_1,A_2,B_2$, respectively, and each is nonnegative.
On $\M$, both effective layers are invertible, so
$\tW_2\T\Sigeps\tW_2$ and $\tW_1\Sigx\tW_1\T$ are positive definite.
The first term can therefore vanish only if
$\Tr(B_1\Sigx B_1\T)=0$, which forces $B_1=0$.
The remaining terms similarly force $A_1=0$, $B_2=0$, and $A_2=0$.
Conversely, setting all four factors to zero gives
$\tW_1=\tW_2=I$ and $S=0$, attaining the lower bound.
\end{proof}

\subsection{Additional experiments with ReLU networks}\label{app:factored-experiments}

We also test whether residual branches shrink when we introduce ReLU
activations. Alongside the linear networks of Section~\ref{sec:factored},
we train two models with two residual blocks each:
\[
    h_0=x,\qquad
    h_\ell=h_{\ell-1}+\mathcal R_\ell(h_{\ell-1})\quad(\ell=1,2),
    \qquad f(x)=h_2.
\]
The residual branches are
\[
    \mathcal R_\ell(h)=
    \begin{cases}
        W_\ell\mathrm{ReLU}(h), & \text{direct model},\\
        A_\ell\mathrm{ReLU}(B_\ell h), & \text{factored model}.
    \end{cases}
\]
All matrices are $d\times d$, and ReLU acts elementwise.
Note that for the direct model, we place the trainable matrix after ReLU so that the residual output can have either sign.\footnote{With blocks $h\mapsto h+\mathrm{ReLU}(W_\ell h)$,
each residual contribution is coordinatewise nonnegative. Their sum can
vanish only if every contribution vanishes, so implementing the identity
would already require zero branch outputs. A matrix after ReLU permits
nonzero residual contributions to cancel across blocks.}

\paragraph{Initialization.}
We draw a Gaussian matrix $R_1$ for each run and rescale it to $\|R_1\|_F=1$, then set
$R_2=(I+R_1)^{-1}-I$. The direct model starts at $W_\ell=R_\ell$.
For the factored model, an SVD $R_\ell=P_\ell D_\ell Q_\ell\T$ gives
\[
    A_\ell=P_\ell D_\ell^{1/2},
    \qquad B_\ell=D_\ell^{1/2}Q_\ell\T.
\]
For linear blocks, both models therefore start with identical residual
maps and implement the identity. We train for $8\times10^5$ steps with $d=16$, $\eta=0.02$,
$B=64$, no weight decay, and six runs per noise setting.
The ReLU models use the same initial matrices. Their matrix products
agree initially, but their functions need not agree or implement the identity.

\paragraph{Measurements.}
For linear blocks, and as we report in Figure~\ref{fig:flip}, we can measure
$\sum_\ell\|W_\ell\|_F$ in the direct model and
$\sum_\ell\|A_\ell B_\ell\|_F$ in the factored model.
These are norms of the matrices implementing the residual maps, rather
than norms of the individual factors $A_\ell$ and $B_\ell$.

For ReLU blocks, we measure the RMS size of each residual branch's
contribution, relative to the input RMS, and sum over the two blocks:
\[
    \mathcal R:=\sum_{\ell=1}^2
      \frac{\big(\E_x\|\mathcal R_\ell(h_{\ell-1})\|^2\big)^{1/2}}
           {\big(\E_x\|x\|^2\big)^{1/2}}.
\]
A small $\mathcal R$ means that the residual branches contribute little
to the network's output. We also check whether the network learns the
identity by measuring its relative RMS function error:
\[
    \mathcal E:=
      \frac{\big(\E_x\|f(x)-x\|^2\big)^{1/2}}
           {\big(\E_x\|x\|^2\big)^{1/2}}.
\]
For this appendix section, we use the same measurements for the linear networks, for comparison.

\begin{table}[htbp]\centering\small
\begin{tabular}{lllcc}
\toprule
parametrization & activation & noise & relative branch RMS & relative function error \\
\midrule
direct & linear & iso & $0.9146$ & $0.0080$ \\
direct & ReLU & iso & $0.8677$ & $0.0087$ \\
direct & linear & aniso & $1.2027$ & $0.0080$ \\
direct & ReLU & aniso & $0.8049$ & $0.0093$ \\
\midrule
factored & linear & iso & $0.0016$ & $0.0014$ \\
factored & ReLU & iso & $0.0028$ & $0.0022$ \\
factored & linear & aniso & $0.0016$ & $0.0014$ \\
factored & ReLU & aniso & $0.0030$ & $0.0023$ \\
\bottomrule
\end{tabular}

\caption{\textbf{Residual branches after training from matched matrix products.}
The branch RMS and function error columns report $\mathcal R$ and
$\mathcal E$, respectively. Entries are means over six runs,
using weights averaged over the final $1\%$ of $8\times10^5$ SGD steps.
Each run is evaluated on 8192 Gaussian inputs independent of training.
All runs use $d=16$, $\eta=0.02$, $B=64$, and $\gamma=0$.}
\label{tab:factored}
\end{table}

\begin{figure}[htbp]
    \centering
    \includegraphics[width=0.95\linewidth]{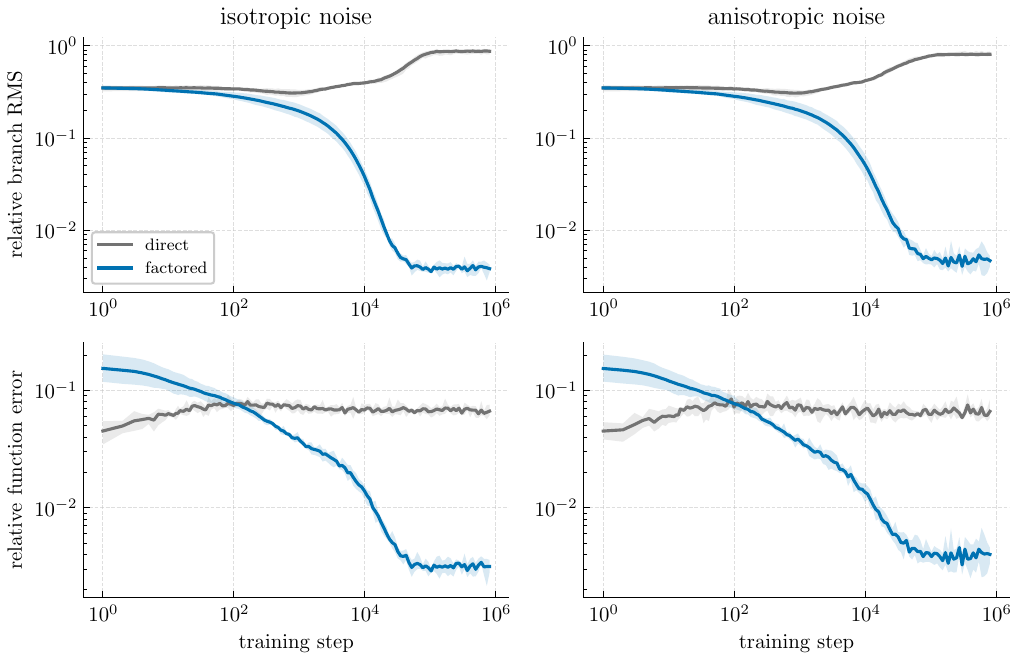}
    \caption{\textbf{Functional measurements of ReLU branch collapse.}
    Top panels show relative branch RMS $\mathcal R$; bottom panels
    show clean function error $\mathcal E$. Columns correspond to isotropic
    and anisotropic label noise. Gray curves show the direct model and blue
    curves the factored model. Lines are means across six runs and shaded
    bands show their ranges.
    At each plotted training step, we evaluate the current weights on
    2048 Gaussian inputs per run, independent of training, without averaging
    weights across steps.
    Training uses $d=16$, $\eta=0.02$, $B=64$, $\gamma=0$, and $8\times10^5$ steps.}
    \label{fig:relu-functional}
\end{figure}

\paragraph{Results.}
We observe similar behavior in the ReLU and linear networks: residual
branch contributions collapse to nearly zero with the factored
parametrization, while they remain nonzero with the direct parametrization
(Table~\ref{tab:factored}, Figure~\ref{fig:relu-functional}).

\FloatBarrier
\section{General covariances}\label{app:diag}

\subsection{Diagonal noise covariance with isotropic inputs}\label{app:diagonalize}

With isotropic inputs, assuming diagonal noise covariance is only a
choice of coordinates. Write $\Sigeps=Q\Lambda Q\T$, where $Q$ is
orthogonal and $\Lambda$ is diagonal, and set
\[
    x'=Q\T x,\qquad y'=Q\T y,\qquad W_\ell'=Q\T W_\ell Q.
\]
Then $\Sigx'=I$, $\Sigeps'=\Lambda$, and $f'(x')=Q\T f(x)$.
Orthogonal transformations preserve Euclidean norms, so the squared loss
and the residual-weight penalty are unchanged. For corresponding
minibatches, the gradients satisfy
\[
    \nabla_{W_\ell'}\ell_{\mathcal B}'
    =Q\T(\nabla_{W_\ell}\ell_{\mathcal B})Q.
\]
Thus rotating the initialization and every minibatch rotates the entire
SGD trajectory.

\subsection{Entropic loss analysis for general covariances}\label{app:general-covariance}

Now let $\Sigx,\Sigeps\succ0$, without assuming they commute, and set
$\gamma=0$. On the identity manifold, write
$G=\tW_1\T\tW_1\succ0$ and $\tW_2=\tW_1^{-1}$.
The gradient norms in Appendix~\ref{app:gradient-norms} give
\begin{equation}
    S(G)=\frac{\eta}{B}\left[
        \Tr(\Sigx)\Tr(\Sigeps G^{-1})
        +\Tr(\Sigeps)\Tr(\Sigx G)\right].
    \label{eq:general-covariance-S}
\end{equation}
The same argument as in Appendix~\ref{app:gram-derivative} gives strict
convexity in $G$. Setting the derivative to zero yields
\[
    -\Tr(\Sigx)G^{-1}\Sigeps G^{-1}
    +\Tr(\Sigeps)\Sigx=0,
    \qquad
    G\Sigx G=c\Sigeps,\quad c:=\frac{\Tr\Sigx}{\Tr\Sigeps}.
\]
Multiplying the latter equation on both sides by $\Sigx^{1/2}$ gives
$(\Sigx^{1/2}G\Sigx^{1/2})^2=c\Sigx^{1/2}\Sigeps\Sigx^{1/2}$.
Taking the positive definite square root therefore gives the unique minimizer
\begin{equation}
    G^\star=\Sigx^{-1/2}
    \left(c\Sigx^{1/2}\Sigeps\Sigx^{1/2}\right)^{1/2}
    \Sigx^{-1/2}.
    \label{eq:general-covariance-Gram}
\end{equation}

If the covariances commute, they share an orthonormal eigenbasis.
Writing their eigenvalues in this basis as $x_i$ and $e_i$, respectively,
the singular values of $\tW_1$ at the minimizer are
$s_i=(c e_i/x_i)^{1/4}$.
For $\Sigx=I$ and $\Tr\Sigeps=d$, this recovers the body's fourth-root law.

\subsection{Testing noncommuting covariances}\label{app:noncommuting}

We compare SGD with the predicted Gram matrix $G^\star$ in two dimensions,
varying the relative orientation of the covariances while keeping their
eigenvalues fixed. With $R_\theta$ a planar rotation, set
\[
    \Sigeps=\frac{2}{1+\kappa}\diag(\kappa,1),\qquad
    \Sigx=\frac{2}{1+\kappa}R_\theta\diag(1,\kappa)R_\theta\T.
\]
Both covariances have trace $2$ and condition number
$\kappa\in\{3,5,10\}$. For each $\kappa$, we compare
$\theta=0^\circ$ and $90^\circ$ (commuting) with $45^\circ$ (noncommuting).

\begin{figure}[t]
    \centering
    \includegraphics[width=0.94\linewidth]{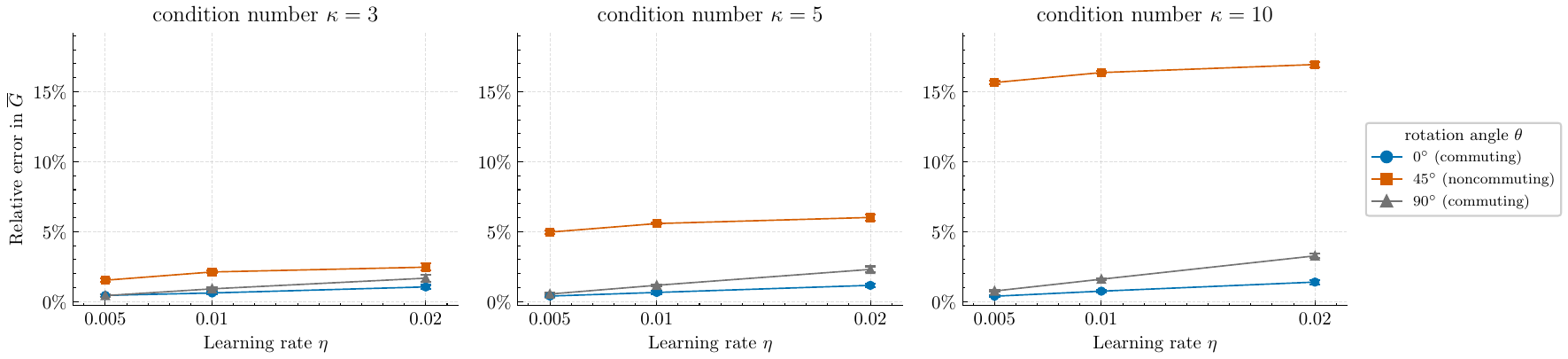}
    \caption{\textbf{Covariance orientation and the accuracy of the entropic
    prediction.} Each panel fixes the covariance eigenvalues and varies
    their relative orientation. Blue circles and gray triangles show the
    commuting controls ($0^\circ$ and $90^\circ$); orange squares show the
    noncommuting case ($45^\circ$).
    Points show $\|\overline G-G^\star\|_F/\|G^\star\|_F$, averaged
    across eight runs, with one standard error; lines connect measurements.
    Here $\overline G$ averages $G=\tW_1\T\tW_1$ sampled at regular intervals
    over the final quarter of each run, and $G^\star$ is computed for each covariance pair.
    All cases start at $\tW_1=\tW_2=I$.
    Training uses batch size $B=8$, and no weight decay.
    Learning rates $0.02$, $0.01$, and $0.005$ use $400{,}000$,
    $1{,}600{,}000$, and $6{,}400{,}000$ steps, respectively, keeping
    $\eta^2t/B=20$ at the endpoint.}
    \label{fig:covariance-prediction}
\end{figure}

\FloatBarrier
Figure~\ref{fig:covariance-prediction} compares the learned Gram matrices
with $G^\star$ for each covariance pair. The noncommuting cases show
larger discrepancies, which remain at smaller learning rates and grow
across the three tested condition numbers. Thus covariance commutativity seems to matter for the accuracy of the entropic prediction in this example.
Explaining this discrepancy, and more generally determining when the predictions derived from entropic loss coincide with the empirical behavior of SGD, remain questions for future work.

\FloatBarrier
\section{Dependence on initialization}\label{app:detsector}

\subsection{Determinant sign and the identity manifold}\label{app:determinant-components}

The two branches of the identity hyperbola $\tw_2\tw_1=1$ are
distinguished by the sign of $\tw_1$. On the positive branch both effective weights
are positive; on the negative branch both are negative.
In Figure~\ref{fig:oned}, different initializations lead runs toward
different branches, along which they move toward $(1,1)$ or $(-1,-1)$,
respectively.

These two branches generalize to two connected components of the identity manifold in higher dimensions.
Each factorization has the form $(\tW_1,\tW_1^{-1})$, and the components are distinguished by the sign of $\det\tW_1$, just as the scalar branches are distinguished by the sign of $\tw_1$.
A continuous path on the manifold can join any two factorizations with the same determinant sign, but cannot join factorizations with opposite signs: changing sign would require passing through a singular $\tW_1$.
As $\tW_1$ approaches singularity on the manifold, its inverse $\tW_2=\tW_1^{-1}$ grows without bound.
In our experiments, different initializations lead to different determinant components.
Note that the component reached need not match the initial determinant sign: the layers generally start away from the identity manifold, where they need not be inverses and their determinant signs can change.

\clearpage
\subsection{Effect of initialization scale}\label{app:orthogonal-freedom}

We next examine how initialization scale changes the initial eigenvalue distribution, before comparing the eigenvalues after training.

\begin{figure}[!htbp]
    \centering
    \includegraphics[width=0.98\linewidth]{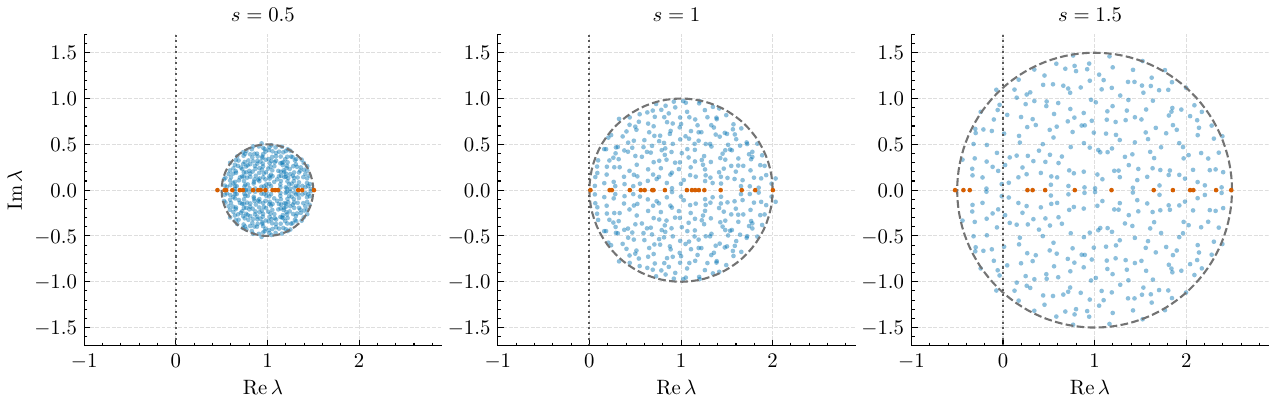}
    \caption{\textbf{Eigenvalue distribution at initialization.}
    Each panel shows one draw of $\tW_1=I+(s/\sqrt d)G$ at $d=400$,
    where the entries of $G$ are independent standard Gaussians.
    Dashed circles have center $1$ and radius $s$; the dotted vertical
    line marks zero real part. Blue points are nonreal eigenvalues and
    orange points are real eigenvalues. The disk reaches the origin at
    $s=1$ and extends into the left half-plane for $s>1$.}
    \label{fig:sectors-init-spectrum}
\end{figure}

For Gaussian initialization, write
\[
    \tW_1=I+\frac{s}{\sqrt d}G,
    \qquad G_{ij}\sim\mathcal N(0,1)\ \text{independently}.
\]
The circular law describes the eigenvalues of $G/\sqrt d$ at large $d$
as approximately uniform over the unit disk \citep{tao2008circular}.
Scaling by $s$ and adding $I$ gives a disk of radius $s$ centered at $1$
(Figure~\ref{fig:sectors-init-spectrum}). For $s<1$, the disk lies in the
right half-plane; for $s>1$, it extends into the left half-plane.
The initial determinant is negative exactly when an odd number of real eigenvalues are negative (nonreal eigenvalues occur in conjugate pairs and thus together contribute a positive product, $z\bar z=|z|^2>0$, to the determinant).
Small initializations ($s<1$) therefore tend to have positive initial determinants.
Once the disk extends past the origin ($s>1$), it includes part of the negative real axis, and the parity of the negative real eigenvalues determines the sign.

\clearpage
\subsection{Learned eigenvalues}\label{app:learned-eigenvalues}

\begin{figure}[!htbp]
    \centering
    \includegraphics[width=0.98\linewidth]{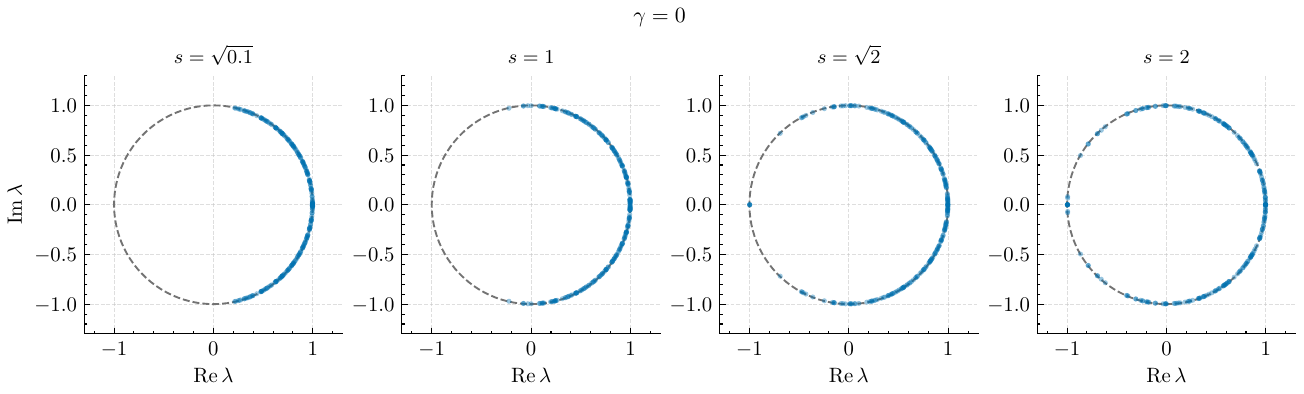}\\[-0.5em]
    \includegraphics[width=0.98\linewidth]{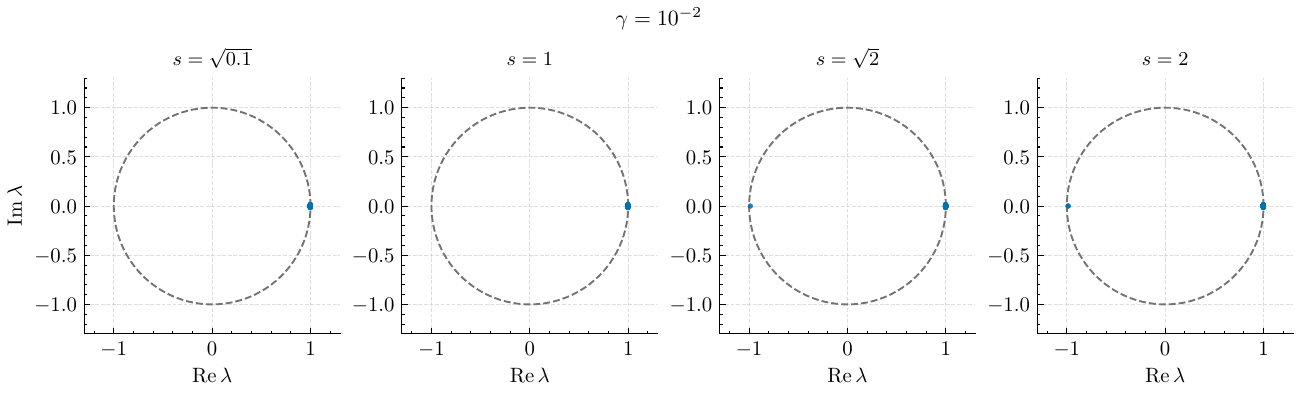}
    \caption{\textbf{Learned eigenvalues across initialization scales.} Each panel overlays the
    eigenvalues of $\tW_1$ from $16$ runs with $d=16$ and isotropic noise.
    The dashed curve is the unit circle. Initial entries of each $W_\ell$
    are independent $\mathcal N(0,s^2/d)$, with
    $s\in\{\sqrt{0.1},1,\sqrt{2},2\}$ from left to right; the second column uses the
    initialization of Figure~\ref{fig:eig}(b). The rows share initial weights
    and use $\gamma=0$ (top) and $\gamma=10^{-2}$ (bottom).
    We train for $2\times10^6$ steps with $\eta=0.02$, $B=64$, then average weights over the final $1\%$ of training.}
    \label{fig:sectors-eigs}
\end{figure}

Figure~\ref{fig:sectors-eigs} compares the learned eigenvalues across initialization scales. Without weight decay (top), they lie near the unit circle, consistent with Section~\ref{sec:iso}. Smaller initializations give an arc in the right half-plane, while larger ones spread the eigenvalues around more of the circle. Figure~\ref{fig:eig}(b) shows the $s=1$ case.

With weight decay (bottom), runs in the positive determinant component
move toward the identity, with all eigenvalues near $+1$.
Larger initializations also produce runs in the negative component.
Within the orthogonal family, weight decay favors a hyperplane reflection
in this component: one eigenvalue is $-1$ and all others are $+1$.

\clearpage
\section{Additional experiments}
\label{app:additional-experiments}

\subsection{Sensitivity to learning rate and batch size}
\label{app:lr-batch-sensitivity}

We examine whether the predicted Gram matrix continues to describe the
solutions learned by SGD across different learning rates and batch sizes.
We use the two-layer setting with isotropic inputs, anisotropic label noise,
and no weight decay, as in Section~\ref{sec:aniso}.
We vary the learning rate over $\eta\in\{0.0025,0.005,0.01\}$ and the batch
size over $B\in\{8,16,32\}$, with ten runs per configuration.

\begin{figure}[htbp]
    \centering
    \includegraphics[width=0.98\linewidth]{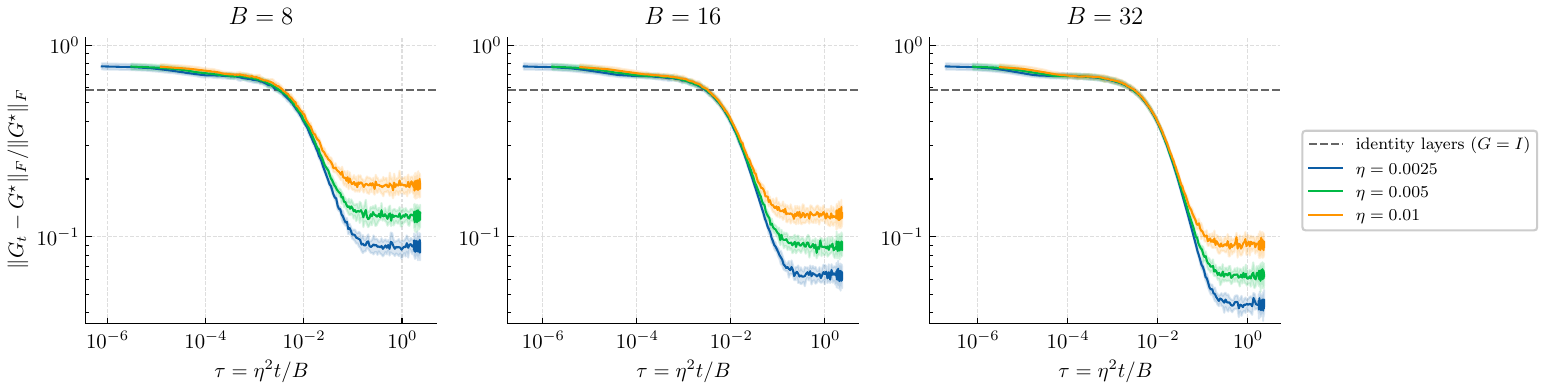}
    \caption{\textbf{Sensitivity to learning rate and batch size.}
    Relative error of the current Gram matrix
    $G_t=\tW_1(t)^\top\tW_1(t)$ against the prediction
    $G^\star=\Sigeps^{1/2}$.
    Curves show means over ten runs; shaded bands show one standard deviation.
    The dashed line is the identity-layer reference $G=I$.
    Here $d=16$ and the diagonal label-noise variances are logarithmically
    spaced with condition number $100$, normalized to sum to $16$.
    Initial residual entries are independent $\mathcal N(0,0.3^2/d)$.}
    \label{fig:lr-batch-trajectories}
\end{figure}

\begin{figure}[!htbp]
    \centering
    \includegraphics[width=0.42\linewidth]{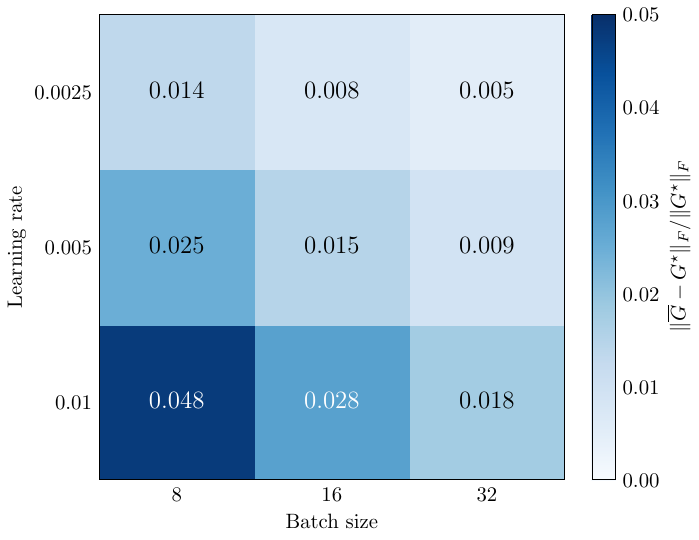}
    \caption{\textbf{Agreement of time-averaged Gram matrices with prediction.}
    For each run in Figure~\ref{fig:lr-batch-trajectories}, we compute
    $\overline G$ by averaging the Gram matrices of checkpoints over the final quarter of training.
    Cells show $\|\overline G-G^\star\|_F/\|G^\star\|_F$, averaged over ten runs.
    The identity-layer reference $G=I$ has relative error $0.585$.}
    \label{fig:lr-batch-heatmap}
\end{figure}

Across all nine configurations, the learned Gram matrices approach the
prediction and continue to fluctuate around it
(Figure~\ref{fig:lr-batch-trajectories}).
Smaller learning rates and larger batches reduce these fluctuations and
improve agreement with the prediction.
Averaging each run's Gram matrices over the final quarter of training gives
relative errors between $0.005$ and $0.048$, averaged across runs
(Figure~\ref{fig:lr-batch-heatmap}).
For comparison, the identity-layer solution $G=I$ has relative error $0.585$.

\clearpage
\subsection{Small and zero label-noise eigenvalues}
\label{app:small-noise-eigenvalues}

We examine how SGD behaves when one label-noise eigenvalue is small or zero.
We use two-dimensional, two-layer linear residual networks with isotropic inputs,
no weight decay, and label-noise covariance $\Sigeps=\diag(\lambda,2-\lambda)$.
We vary $\lambda\in\{10^{-1},10^{-2},10^{-3},10^{-4},0\}$ and compare learning rates
$\eta\in\{0.001,0.0005\}$, with ten runs per configuration.

\begin{figure}[htbp]
    \centering
    \includegraphics[width=0.98\linewidth]{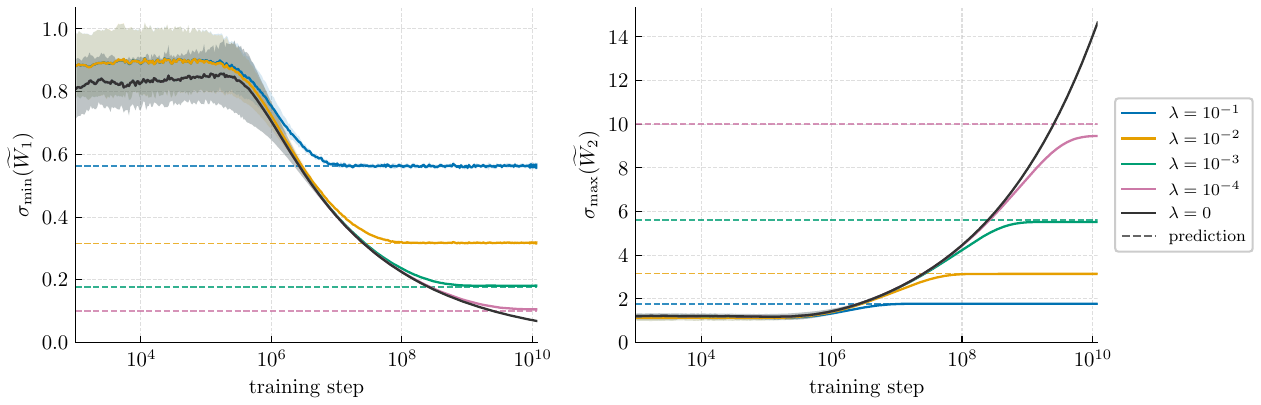}
    \caption{\textbf{Small and zero label-noise eigenvalues.}
    Smallest singular value of $\tW_1$ (left) and largest singular value of
    $\tW_2$ (right), for $\eta=0.0005$.
    Curves show means over ten runs; shaded bands show one standard deviation.
    Dashed lines show $\lambda^{1/4}$ and $\lambda^{-1/4}$ for positive $\lambda$;
    the zero case has no finite on-manifold minimizer.
    All runs use batch size $B=1$ and independent initial residual weight entries
    drawn from $\mathcal N(0,0.3^2/2)$.}
    \label{fig:small-noise-trajectories}
\end{figure}

\begin{figure}[htbp]
    \centering
    \includegraphics[width=0.98\linewidth]{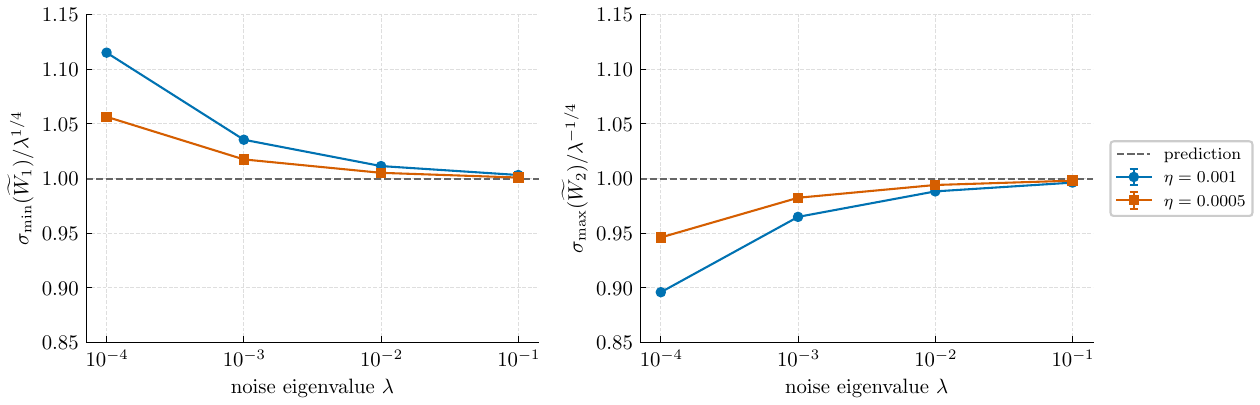}
    \caption{\textbf{Agreement with the predicted singular values at two learning rates.}
    For each run, we average singular values over uniformly
    sampled checkpoints in the final $10\%$ of training; points show means
    across ten runs and error bars show one standard deviation.
    Other settings match Figure~\ref{fig:small-noise-trajectories}.
    The zero-eigenvalue case is omitted because it has no finite prediction
    by which to normalize.}
    \label{fig:small-noise-rate-check}
\end{figure}

Smaller noise eigenvalues lead to stronger contraction in the first effective
layer and expansion in the second (Figure~\ref{fig:small-noise-trajectories}).
For positive $\lambda$, the learned singular values settle near the predictions
$\lambda^{1/4}$ and $\lambda^{-1/4}$; the remaining discrepancies are reduced
at smaller learning rate (Figure~\ref{fig:small-noise-rate-check}).
At $\lambda=0$, the entropic objective has no finite minimizer on the identity
manifold: its infimum is approached as the first layer becomes singular and
the second grows without bound.

\clearpage
\end{document}